\documentclass[11 pt]{elsarticle}

\usepackage{framed,multirow}
\usepackage[utf8]{inputenc}
\usepackage[T1]{fontenc}
\usepackage{siunitx}
\usepackage{graphicx}

\usepackage{hyperref}
\usepackage[ruled,vlined]{algorithm2e}

\usepackage{xcolor}

\SetKwInOut{KwRequire}{Require}
\SetKwInOut{KwEnsure}{Ensure}
\SetKwComment{Comment}{$\triangleright$\ }{}

\newcounter{algline}

\DontPrintSemicolon

\usepackage{subcaption}
\usepackage{hyperref,float}
\usepackage{booktabs}
\usepackage{makecell}
\usepackage{tabu}
\usepackage{float}
\usepackage{placeins}
\usepackage{multirow} 
\usepackage{amssymb}
\usepackage{latexsym}
\usepackage{amsmath}
\usepackage[
left=3.8cm,
right=3.8cm,
top=4.8cm,
bottom=4.8cm
]{geometry}

\usepackage{caption}
\usepackage{amsthm}

\theoremstyle{plain}

\theoremstyle{definition}

\theoremstyle{remark}

\usepackage{amsfonts}
\usepackage{float}
\usepackage{url}
\usepackage{xcolor}

\definecolor{newcolor}{rgb}{.8,.349,.1}

\makeatletter
\def\ps@pprintTitle{%
  \let\@oddhead\@empty
  \let\@evenhead\@empty
  \def\@oddfoot{%
    \reset@font
    \small
    \parbox[t]{\textwidth}{%
      \raggedright
      Preprint submitted to \textit{Machine Learning with Applications}%
      \hfill
      \today
    }%
  }%
  \let\@evenfoot\@oddfoot
}
\makeatother

\usepackage[nolist]{acronym}
\begin{acronym}
	\acro{PDE}[PDE]{partial differential equation}
	\acro{FE}[FE]{finite element}
	\acro{FEM}[FEM]{\ac{FE} method}
	\acro{RB}[RB]{reduced basis}
	\acro{CRB}[CRB]{certified \ac{RB}}
	\acro{EIM}[EIM]{empirical interpolation method}
	\acro{SCM}[SCM]{successive constraint method}
	\acro{SCRBE}[SCRBE]{static condensation reduced basis element}
	\acro{SC}[SC]{static condensation}
	\acro{PR}[PR]{port reduction}
	\acro{PR-SCRBE}[PR-SCRBE]{\ac{PR}-\ac{SCRBE}}
	\acro{PR-RB}[PR-RB]{port-reduced \ac{RB}}
	\acro{DOF}{degrees of freedom}
	\acro{ML}{machine learning}
	\acro{POD}{proper orthogonal decomposition}
	\acro{ROM}{reduced-order modelling}
	\acro{RSM}{response surface methodology}
	\acro{CAD}{computer-aided design}
	\acro{MA}{manifold-alignment}
	\acro{CRM}{common research model}
	\acro{RSM}{response surface methodology}
	\acro{NN}{neural network}
	\acro{RBF}{radial basis function}
	\acro{MA}{manifold-alignment}
	\acro{AC}{archetype component}
	\acro{IC}{instantiated component}
	\acro{MDO}{multidisciplinary design optimization}
	\acro{MUMPS}{MUltifrontal Massively Parallel sparse direct Solver}
    \acro{GPR}{Gaussian process regression}
    \acro{GP}{Gaussian process}
    \acro{QoIs}{quantities of interest}
\end{acronym}

\journal{arXiv}
\date{\today}

\begin{document}


\begin{frontmatter}


\title{
Symbolic Machine Learning for Vapor--Liquid Equilibrium Prediction in $C_x$--$N_2$ Binary Mixtures
}


\author[label1]{Bongseok Kim}

\author[label3]{Suman Chakraborty} 

\author[label3]{Gary Huang}

\author[label3]{Mehek Mathur}

\author[label1,label2]{Guang Lin}

\author[label3]{Li Qiao\corref{cor2}}
\cortext[cor2]{Corresponding author}
\ead{lqiao@purdue.edu}

\affiliation[label1]{organization={Purdue University, School of Mechanical Engineering},
            postcode={47906}, 
            state={IN},
            country={United States}}

\affiliation[label2]{organization={Purdue University, Department of Mathematics},
            postcode={47906},
            state={IN},
            country={United States}}

\affiliation[label3]{organization={Purdue University, School of Aeronautics and Astronautics},
            postcode={47906},
            state={IN},
            country={United States}}

\begin{abstract}

Accurate prediction of vapor--liquid equilibrium (VLE) for hydrocarbon--nitrogen mixtures remains challenging for cubic equations of state, particularly across broad ranges of composition and hydrocarbon chain length. While deep learning models can provide accurate predictions, they often lack interpretability and explicit analytical expressions. 
In this work, we propose a symbolic machine learning approach to discover interpretable symbolic corrections to Peng--Robinson equation-of-state (PR-EOS) predictions from experimental data. The proposed approach adopts a two-level strategy: symbolic expressions are first identified for individual hydrocarbon systems, after which their coefficients are represented as functions of carbon number to enable accurate prediction across different hydrocarbon systems. 
The results demonstrate significantly improved prediction accuracy over the original PR-EOS across all hydrocarbon--nitrogen systems.
Overall, the proposed approach provides an interpretable symbolic correction framework for improving PR-EOS predictions of hydrocarbon--nitrogen VLE.

\end{abstract}
\end{frontmatter}


\section{Introduction}

Accurate thermodynamic modeling of multicomponent hydrocarbon mixtures is essential in many engineering applications, including high-pressure fuel injection~\cite{Qiu2015} and supercritical combustion~\cite{Reitz1987,Wang2018}.
Reliable vapor--liquid equilibrium (VLE) prediction is particularly important because phase behavior strongly influences fuel atomization, transport processes, and thermodynamic efficiency \cite{ReitzBeale1999,QiaoHighPressure}. 
To predict VLE, equations of state (EOS) have become one of the most widely used thermodynamic models for describing pressure--temperature--composition relationships in multicomponent mixtures.

Among cubic equations of state, the Peng--Robinson (PR)-EOS~\cite{PengRobinson1976} has become one of the most widely used thermodynamic models for predicting vapor--liquid equilibria as well as volumetric and thermodynamic properties of pure substances and mixtures. 
Extensive efforts have been devoted to improving the PR-EOS through modifications of the attraction term, volume translation, additional model terms, and mixing rules~\cite{Lopez2017}.
Despite its widespread use, the classical PR-EOS still exhibits non-negligible prediction errors for hydrocarbon mixtures, particularly under high-pressure and near-critical conditions where complex intermolecular interactions become increasingly important~\cite{Lopez2017,Privat2013}.

Considerable effort has therefore been devoted to improving the predictive capability of the PR-EOS. One important direction is the development of predictive group-contribution mixing rules for hydrocarbon mixtures~\cite{Privat2008a,Privat2008b}. Other studies have proposed generalized correlations for estimating binary interaction parameters over broad thermodynamic conditions~\cite{Fateen2013,Mohammed2018,Abudour2014}. In parallel, extensive experimental measurements have been reported for hydrocarbon--nitrogen systems. Representative VLE datasets include nitrogen--decane~\cite{GarciaSanchez2009}, nitrogen--dodecane~\cite{GarciaCordova2011}, and nitrogen--hydrocarbon systems at elevated pressures~\cite{Azarnoosh1963,Llave1988}. High-pressure solubility measurements have also been reported for heavy n-alkanes and other hydrocarbon systems~\cite{DAvila1976,Prausnitz1959,Pearce1993,Tong1999,Gao1999}. These experimental datasets provide valuable benchmarks for validating thermodynamic models.

Recently, machine learning has emerged as an alternative framework for thermodynamic modeling. Artificial neural networks have been applied to vapor--liquid equilibrium (VLE) prediction~\cite{Petersen1994,Sharma1999}. Subsequent studies extended neural-network-based VLE prediction to a wider range of binary systems, including refrigerant mixtures~\cite{Mohanty2005,Mohanty2006,Ganguly2003}. 
Neural networks have also been employed for hydrocarbon mixtures, equilibrium K-value prediction, and asymmetric binary systems~\cite{Habiballah1996,Ghanadzadeh2008,Abedini2011, chakraborty2022vapor}. 
While these methods generally improve prediction accuracy, fully data-driven neural-network models do not provide explicit analytical expressions, making the learned thermodynamic relationships difficult to interpret.

To address this limitation, symbolic regression has emerged as an attractive alternative for discovering explicit mathematical expressions directly from data~\cite{Koza1992,Schmidt2009}. Instead of fitting a predefined functional form, symbolic regression simultaneously identifies both the analytical structure and the associated numerical coefficients of the underlying relationships. Consequently, the resulting models remain compact and interpretable while retaining sufficient flexibility to capture nonlinear thermodynamic behavior. 
These characteristics make symbolic regression particularly suitable for constructing interpretable corrections to existing physics-based models rather than replacing them entirely. 
A recent study demonstrated the potential of symbolic regression for improving cubic equations of state through data-driven modifications of EOS parameters for liquid-phase density prediction~\cite{yang2025symbolic}.
In contrast, the present study considers symbolic regression for vapor--liquid equilibrium prediction in multicomponent hydrocarbon--nitrogen mixtures.

In this work, we propose a symbolic machine learning framework for constructing interpretable post-processing corrections to the PR-EOS for hydrocarbon--nitrogen vapor--liquid equilibrium prediction. Rather than directly replacing the underlying thermodynamic model, the proposed approach learns symbolic correction functions for the residual errors of the PR-EOS, thereby preserving its physical foundation while systematically reducing prediction errors. The framework adopts a two-level learning strategy. First, system-specific symbolic expressions are independently discovered for each hydrocarbon system, allowing the dominant correction mechanisms to be identified without imposing predefined functional forms. Next, recurring symbolic structures shared across different hydrocarbon systems are extracted to construct a common symbolic basis. Finally, only the basis coefficients are parameterized as smooth functions of the carbon number, yielding a unified symbolic correction model that generalizes across multiple hydrocarbon systems while preserving compact analytical expressions and physical interpretability.

The main contributions of this work are summarized as follows:

\begin{enumerate}
\item \textbf{Interpretable symbolic correction of the PR-EOS.}
The proposed approach learns explicit symbolic correction functions for the residual errors of the PR-EOS, preserving the underlying thermodynamic model while improving prediction accuracy.

\item \textbf{Two-level symbolic machine learning.}
System-specific symbolic expressions are first discovered independently, after which shared symbolic basis functions and carbon-number-dependent coefficients are identified to construct a unified symbolic correction model.

\item \textbf{Accurate symbolic VLE prediction.}
The proposed approach consistently improves pressure and vapor-phase composition predictions while maintaining compact closed-form symbolic expressions.
\end{enumerate}

The remainder of this paper is organized as follows. Section~2 presents the proposed multilevel symbolic regression method. Section~3 describes the experimental datasets and presents the numerical results. Finally, Section~4 concludes the paper and discusses future research directions.


\section{Peng--Robinson Equation of State (PR-EOS)}

We begin by reviewing the PR-EOS~\cite{Lopez2017, chakraborty2022vapor}, which serves as the baseline thermodynamic model throughout this work. The proposed symbolic machine learning approach is designed to correct the prediction errors of the PR-EOS rather than replace the underlying thermodynamic formulation. This section therefore briefly summarizes the governing equations of the PR-EOS and the corresponding phase-equilibrium calculations used throughout the remainder of this paper.

For a single-phase mixture, the pressure is expressed as
\begin{equation}
P = \frac{RT}{v - b} - \frac{a(T)}{v(v+b) + b(v-b)}.
\end{equation}

Here, $R$ is the universal gas constant, $T$ is the absolute temperature, and $v$ is the molar volume. The parameter $a(T)$ represents the attractive intermolecular interactions, whereas $b$ accounts for the excluded-volume effect arising from the finite molecular size. These parameters are determined from the critical properties and acentric factors of the individual components.

For each component $i$, the attractive parameter $a_i$ and the covolume parameter $b_i$ are evaluated as
\begin{align}
a_i &= 0.45724 \, \frac{R^2 T_{c,i}^2}{P_{c,i}} \, \alpha_i(T), \\
b_i &= 0.07780 \, \frac{R T_{c,i}}{P_{c,i}}.
\end{align}

The temperature-dependent attraction function is
\begin{equation}
\alpha_i(T)=
\left[
1+\left(
0.37464+1.54226\omega_i-0.26992\omega_i^2
\right)
\left(
1-\sqrt{T/T_{c,i}}
\right)
\right]^2,
\end{equation}
where $T_{c,i}$, $P_{c,i}$, and $\omega_i$ denote the critical temperature, critical pressure, and acentric factor of component $i$, respectively.

For a multicomponent mixture with mole fractions $x_i$, the pure-component parameters are combined through the conventional quadratic mixing rules,
\begin{align}
a &= \sum_i\sum_j x_i x_j a_{ij}, \\
b &= \sum_i x_i b_i,
\end{align}
where
\begin{equation}
a_{ij}=(1-k_{ij})\sqrt{a_i a_j}.
\end{equation}

Here, $k_{ij}$ denotes the binary interaction parameter, which accounts for non-ideal interactions between unlike species. It satisfies the symmetry condition $k_{ij}=k_{ji}$ together with $k_{ii}=0$.

Introducing the dimensionless parameters
\begin{equation}
A=\frac{aP}{R^2T^2},
\qquad
B=\frac{bP}{RT},
\qquad
Z=\frac{Pv}{RT},
\end{equation}
the compressibility factor $Z$ is obtained by solving the cubic equation
\begin{equation}
Z^3-(1-B)Z^2+(A-3B^2-2B)Z-(AB-B^2-B^3)=0.
\end{equation}

The physically appropriate real root is selected according to the phase of interest.

Phase equilibrium calculations are subsequently performed through fugacity equality between the liquid and vapor phases. Within the PR-EOS, the fugacity coefficient of component $i$ is evaluated as
\begin{align}
\ln\phi_i
={}&
\frac{b_i}{b}(Z-1)-\ln(Z-B)
\nonumber\\
&
-\frac{A}{2\sqrt{2}B}
\left(
\frac{2\sum_j x_j a_{ij}}{a}
-\frac{b_i}{b}
\right)
\ln\left(
\frac{Z+(1+\sqrt2)B}
{Z+(1-\sqrt2)B}
\right).
\end{align}

The fugacity coefficients provide the thermodynamic equilibrium criterion for each component and are therefore the key quantities for vapor--liquid equilibrium calculations.

In practical VLE calculations, the computational procedure consists of the following steps:

\begin{enumerate}
\item Evaluate the pure-component parameters $a_i$, $b_i$, and $\alpha_i(T)$.
\item Construct the mixture parameters $a$ and $b$ using the quadratic mixing rules.
\item Solve the cubic equation of state to obtain the compressibility factor $Z$.
\item Compute the fugacity coefficients for each component in the liquid and vapor phases.
\item Determine the equilibrium pressure and vapor composition by satisfying the phase-equilibrium conditions.
\end{enumerate}

\section{Symbolic regression for correction of PR-EOS}
\subsection{System-specific symbolic regression}

\begin{figure}[htp!]
\centering
\begin{subfigure}[b]{0.85\linewidth}
    \centering
    \includegraphics[width=\linewidth]{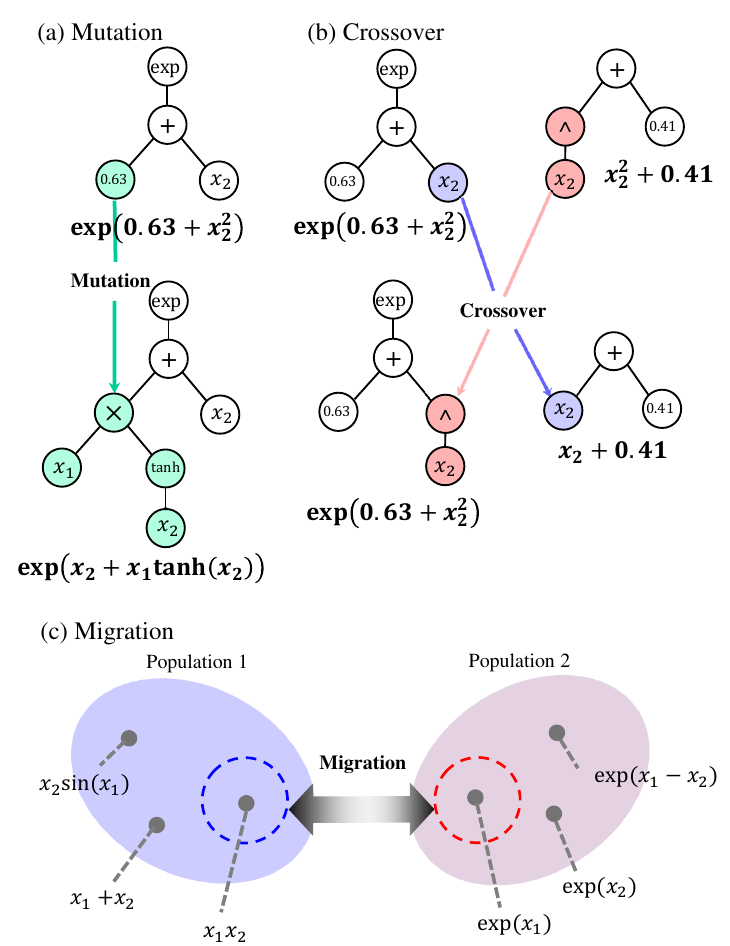}
\end{subfigure}
\caption{
Illustration of the evolutionary search process in symbolic regression.
(a) Mutation replaces a randomly selected subtree with a newly generated symbolic subtree, modifying the expression from $\exp(0.63+x_2)$ to $\exp(x_1\tanh(x_2)+x_2)$.
(b) Crossover exchanges randomly selected subtrees between two parent expressions to generate new candidate expressions.
(c) Migration transfers symbolic expressions between independent populations during parallel evolution.
After structural updates, symbolic simplification and numerical constant optimization are subsequently applied before evaluating the updated expressions.
}
\label{fig:pysr_evolutionary_search}
\end{figure}

Symbolic regression aims to discover explicit mathematical expressions directly from data by simultaneously identifying both the analytical structure and the associated numerical coefficients. In this work, symbolic regression is performed using the evolutionary search algorithm implemented in PySR~\cite{Cranmer2023}, as illustrated in Fig.~\ref{fig:pysr_evolutionary_search}. During each generation, candidate symbolic expressions are modified through three complementary genetic operators. First, mutation introduces structural diversity by replacing a randomly selected subtree with a newly generated symbolic subtree, thereby exploring new functional forms while preserving the remaining expression structure, as shown in Fig.~\ref{fig:pysr_evolutionary_search}a. Second, crossover combines information from two parent expressions by exchanging randomly selected subtrees, enabling the construction of offspring that inherit advantageous components from both parents, as shown in Fig.~\ref{fig:pysr_evolutionary_search}b. Third, migration periodically transfers high-quality expressions between independent populations, promoting information sharing while maintaining diversity across parallel evolutionary searches, as shown in Fig.~\ref{fig:pysr_evolutionary_search}c. After these structural modifications, each candidate expression is algebraically simplified and its numerical constants are refined through numerical optimization before its fitness is evaluated. This iterative process jointly explores the symbolic search space while progressively improving the predictive accuracy of the discovered expressions.

\begin{table}[htp!]
\centering
\small
\caption{Summary of the experimental high-pressure VLE datasets for binary N$_2$ + n-alkane systems used in this study.}
\label{tab:exp_datasets}
\renewcommand{\arraystretch}{1.2}
\begin{tabular}{lcccc}
\hline
System & Source & Temperature levels (K) & $P_{\max}$ (MPa) & Data points \\
\hline
N$_2$ + n-C$_5$H$_{12}$ &
\cite{SilvaOliver2006} &
\makecell[l]{344.3, 377.9, 411.0,\\427.1, 447.9} &
35 &
71 \\

N$_2$ + n-C$_6$H$_{14}$ &
\cite{EliosaJimenez2007} &
\makecell[l]{344.6, 378.2, 411.1,\\444.5, 466.0, 488.4} &
50 &
72 \\

N$_2$ + n-C$_7$H$_{16}$ &
\cite{GarciaSanchez2007} &
\makecell[l]{313.6, 344.3, 377.8, 411.0,\\444.5, 466.0, 488.2,\\502.7, 513.1, 523.7} &
50 &
136 \\

N$_2$ + n-C$_9$H$_{20}$ &
\cite{SilvaOliver2007} &
\makecell[l]{344.3, 411.1, 466.0,\\502.7, 543.4} &
50 &
70 \\

N$_2$ + n-C$_{10}$H$_{22}$ &
\cite{GarciaSanchez2009} &
\makecell[l]{344.3, 377.9, 411.0, 444.5,\\488.2, 523.7, 563.1} &
50 &
142 \\

N$_2$ + n-C$_{12}$H$_{26}$ &
\cite{GarciaCordova2011} &
\makecell[l]{344.4, 377.9, 411.1, 444.6,\\488.2, 523.7, 593.5} &
60 &
169 \\
\hline
\end{tabular}
\end{table}

The experimental datasets employed in this study are summarized in Table~\ref{tab:exp_datasets}. The datasets consist of high-pressure VLE measurements for six binary N$_2$ + n-alkane systems spanning n-pentane to n-dodecane. For each system, measurements were conducted over multiple temperature levels covering a wide range of thermodynamic conditions, providing a total of 660 experimental data points for model development and evaluation.
For each hydrocarbon--nitrogen system, the experimental VLE measurements are used directly to construct the symbolic regression dataset. The reduced temperature \(T_r\) and liquid-phase nitrogen mole fraction \(x_{\mathrm{N}_2}\) are taken as input variables, while the regression targets are defined as the residuals between the experimental measurements and the corresponding PR-EOS predictions. Specifically,
\begin{align}
\Delta P &= P_{\mathrm{exp}}-P_{\mathrm{PR}},\\
\Delta y_{\mathrm{N}_2} &= y_{\mathrm{exp}}-y_{\mathrm{PR}},
\end{align}
where \(P_{\mathrm{exp}}\) and \(y_{\mathrm{exp}}\) denote the measured equilibrium pressure and vapor-phase nitrogen mole fraction, respectively, and \(P_{\mathrm{PR}}\) and \(y_{\mathrm{PR}}\) are the corresponding PR-EOS predictions. Separate symbolic regression models are then identified for the pressure and vapor-composition corrections in each binary system.

Algorithm~\ref{alg:system_symbolic} summarizes the proposed system-specific symbolic learning procedure. Rather than directly learning the equilibrium pressure and vapor composition, symbolic regression is applied to the residual errors of the PR-EOS, allowing the underlying thermodynamic formulation to be retained while correcting its systematic prediction errors. For each hydrocarbon--nitrogen system, separate symbolic regression models are constructed for the pressure and vapor-phase composition residuals because these quantities exhibit distinct nonlinear behaviors. The symbolic expressions are evolved using the procedure described in the previous section, and the final expressions are selected from the Pareto front by balancing prediction accuracy and symbolic complexity. The learned correction functions are then added to the original PR-EOS predictions to obtain the corrected pressure and vapor composition.

\begin{algorithm}[htp!]
\caption{System-specific symbolic regression for PR-EOS correction}
\label{alg:system_symbolic}

\KwRequire{System data $\mathcal{D}^{(s)}=\{(T_r^i,x_{N_2}^i,P_{\mathrm{PR}}^i,y_{\mathrm{PR}}^i,P_{\mathrm{exp}}^i,y_{\mathrm{exp}}^i)\}_{i=1}^{N_s}$.}
\KwEnsure{Corrected predictions $P_{\mathrm{corr}}^{(s)}(T_r,x_{N_2})$ and $y_{\mathrm{corr}}^{(s)}(T_r,x_{N_2})$.}

\textcolor{gray}{$\triangleright$ Residual construction and normalization} \\
Construct
$\Delta P^i=P_{\mathrm{exp}}^i-P_{\mathrm{PR}}^i$
and
$\Delta y^i=y_{\mathrm{exp}}^i-y_{\mathrm{PR}}^i$
for all samples \\
Normalize $(T_r,x_{N_2})$, $\Delta P$, and $\Delta y$ within system $s$\;

\textcolor{gray}{$\triangleright$ Independent symbolic regression} \\
Perform symbolic regression for
$(T_r,x_{N_2}) \mapsto \Delta P$ \\
Perform symbolic regression for
$(T_r,x_{N_2}) \mapsto \Delta y$ \\
Select the final symbolic expressions from the Pareto front according to prediction accuracy and symbolic complexity\;

\textcolor{gray}{$\triangleright$ PR-EOS correction} \\
Construct
$P_{\mathrm{corr}}^{(s)}
=
P_{\mathrm{PR}}
+
\Delta P^{(s)}(T_r,x_{N_2})$
and
$y_{\mathrm{corr}}^{(s)}
=
y_{\mathrm{PR}}
+
\Delta y^{(s)}(T_r,x_{N_2})$ \\
\Return{$P_{\mathrm{corr}}^{(s)}(T_r,x_{N_2})$ and $y_{\mathrm{corr}}^{(s)}(T_r,x_{N_2})$}\;
\end{algorithm}

Algorithm~\ref{alg:multilevel_symbolic} summarizes the proposed multilevel symbolic regression procedure. Rather than maintaining an independent symbolic expression for each hydrocarbon--nitrogen system, the proposed approach constructs a unified symbolic correction model by identifying symbolic structures shared across multiple systems. To this end, the system-specific symbolic expressions obtained in the first stage are first analyzed to extract their constituent symbolic basis functions. Recurring symbolic structures are subsequently identified to construct shared symbolic basis libraries for the pressure and vapor-composition corrections.

For each hydrocarbon system, the correction function is represented as a linear combination of the shared symbolic basis functions,
\begin{equation}
\Delta P^{(s)}(T_r,x_{N_2})
=
\sum_{m=1}^{5}
\alpha_{m}^{(P,s)}
\phi_{m}^{(P)}(T_r,x_{N_2}),
\label{eq:shared_pressure}
\end{equation}
where $\phi_{m}^{(P)}$ denotes the $m$th shared symbolic basis function and $\alpha_{m}^{(P,s)}$ is its corresponding coefficient for system $s$. Likewise, the vapor-composition correction is represented as
\begin{equation}
\Delta y^{(s)}(T_r,x_{N_2})
=
\sum_{m=1}^{6}
\alpha_{m}^{(y,s)}
\phi_{m}^{(y)}(T_r,x_{N_2}),
\label{eq:shared_composition}
\end{equation}

Rather than estimating these coefficients independently for each hydrocarbon system, the proposed approach parameterizes them as functions of the carbon number, allowing the symbolic basis functions to remain common while their coefficients vary systematically with hydrocarbon chain length. The basis-internal constants together with the coefficient functions are then optimized over the merged dataset using gradient-based optimization~\cite{Rumelhart1986}. Finally, the learned symbolic corrections are added to the original PR-EOS predictions to obtain
\begin{equation}
\begin{aligned}
P_{\mathrm{corr}}^{(s)}
&=
P_{\mathrm{PR}}
+
\Delta P^{(s)}(T_r,x_{N_2}),\\
y_{\mathrm{corr}}^{(s)}
&=
y_{\mathrm{PR}}
+
\Delta y^{(s)}(T_r,x_{N_2}),
\end{aligned}
\label{eq:multilevel_reconstruction}
\end{equation}
yielding the final predictions
$P_{\mathrm{corr}}(T_r,x_{N_2},n_C)$
and
$y_{\mathrm{corr}}(T_r,x_{N_2},n_C)$.

\begin{algorithm}[htp!]
\caption{Regression using shared symbolic basis functions}
\label{alg:multilevel_symbolic}

\KwRequire{Merged dataset $\mathcal{D}=\bigcup_s\mathcal{D}^{(s)}$ with system descriptor $n_C$.}
\KwEnsure{Corrected predictions $P_{\mathrm{corr}}(T_r,x_{N_2},n_C)$ and $y_{\mathrm{corr}}(T_r,x_{N_2},n_C)$.}

\textcolor{gray}{$\triangleright$ Shared symbolic basis construction} \\
Extract symbolic basis functions from the system-specific expressions for $\Delta P$ \\
Extract symbolic basis functions from the system-specific expressions for $\Delta y$ \\
Construct the shared basis library $\{\phi_m^{(P)}\}_{m=1}^{M_P}$ for $\Delta P$ \\
Construct the shared basis library $\{\phi_m^{(y)}\}_{m=1}^{M_y}$ for $\Delta y$\;

\textcolor{gray}{$\triangleright$ Independent coefficient learning} \\
Represent
$\Delta P=\sum_{m=1}^{M_P}\alpha_m^{(P)}(n_C)\phi_m^{(P)}(T_r,x_{N_2})$ \\
Represent
$\Delta y=\sum_{m=1}^{M_y}\alpha_m^{(y)}(n_C)\phi_m^{(y)}(T_r,x_{N_2})$ \\
Optimize the basis-internal constants and the coefficient functions
$\alpha_m^{(P)}(n_C)$ for $\Delta P$ using gradient descent \\
Optimize the basis-internal constants and the coefficient functions
$\alpha_m^{(y)}(n_C)$ for $\Delta y$ using gradient descent\;

\textcolor{gray}{$\triangleright$ PR-EOS correction} \\
Construct
$P_{\mathrm{corr}}=P_{\mathrm{PR}}+\Delta P(T_r,x_{N_2},n_C)$ \\
Construct
$y_{\mathrm{corr}}=y_{\mathrm{PR}}+\Delta y(T_r,x_{N_2},n_C)$ \\
\Return{$P_{\mathrm{corr}}(T_r,x_{N_2},n_C)$ and $y_{\mathrm{corr}}(T_r,x_{N_2},n_C)$}\;
\end{algorithm}

\begin{table}[htp!]
\centering
\small
\caption{
System-specific symbolic expressions discovered for the pressure correction term $\Delta P$ in six n-alkane + N$_2$ binary systems. Highlighted subexpressions denote recurring symbolic forms identified across multiple systems. The representative basis column indicates the symbolic basis functions selected for the subsequent multilevel symbolic regression.
}
\label{tab:system_specific_normalized_dp}
\renewcommand{\arraystretch}{1.35}
\begin{tabular}{l c c p{0.60\linewidth}}
\hline
System & \makecell[c]{Relative \\ error [\%]} & \makecell[c]{Representative \\ basis} & Discovered expression for $\Delta P(T_r, x_{N_2})$ \\
\hline

C$_5$H$_{12}$
& $0.020$
& $\phi_1^{(P)}$
& $\displaystyle
\Delta P
=
a_1
\underbrace{
\left(
x_{N_2}+a_2-e^{a_3T_r+x_{N_2}}
\right)
}_{\text{(A) linear--exponential form}}
$ \\

C$_6$H$_{14}$
& $0.039$
& --
& $\displaystyle
\Delta P
=
a_1
\underbrace{
\left(
x_{N_2}+a_2-e^{a_3x_{N_2}}
\right)
}_{\text{(A) linear--exponential form}}
$ \\

C$_7$H$_{16}$
& $0.057$
& $\phi_2^{(P)}$
& $\displaystyle
\Delta P
=
a_1
\underbrace{
\left(
T_r+x_{N_2}^2
\right)
\left(
x_{N_2}+a_2
\right)
}_{\text{(B) coupled polynomial form}}
+a_3
$ \\

C$_9$H$_{20}$
& $0.013$
& $\phi_3^{(P)}$
& $\displaystyle
\Delta P
=
a_1
\underbrace{
\left(
T_r+x_{N_2}+a_2
\right)
\left(
x_{N_2}+a_3
\right)
}_{\text{(B) coupled polynomial form}}
$ \\

C$_{10}$H$_{22}$
& $0.008$
& $\phi_4^{(P)}$
& $\displaystyle
\Delta P
=
\underbrace{
\left(
a_1T_r+a_2x_{N_2}
\right)
}_{\text{(B) coupled polynomial form}}
\,
\underbrace{
\left(
e^{x_{N_2}+a_3(T_r+a_4x_{N_2}^2)}
+a_5
\right)
}_{\text{(C) exponential nonlinear form}}
+a_6
$ \\

C$_{12}$H$_{26}$
& $0.111$
& $\phi_5^{(P)}$
& $\displaystyle
\Delta P
=
\underbrace{
a_2^{-T_r}
}_{\text{(D) inverse-temperature form}}
\,
\underbrace{
\left(
x_{N_2}-e^{x_{N_2}}+a_1
\right)
}_{\text{(A) linear--exponential form}}
$ \\

\hline
\end{tabular}
\end{table}

\begin{table}[htp!]
\centering
\small
\caption{
System-specific symbolic expressions discovered for the vapor-composition correction term $\Delta y$ in six n-alkane + N$_2$ binary systems. Highlighted subexpressions denote recurring symbolic forms identified across multiple systems. The representative basis column indicates the symbolic basis functions selected for the subsequent multilevel symbolic regression.
}
\label{tab:system_specific_normalized_dy}
\renewcommand{\arraystretch}{1.35}
\begin{tabular}{l c c p{0.60\linewidth}}
\hline
System & \makecell[c]{Relative \\ error [\%]} & \makecell[c]{Representative \\ basis} & Discovered expression for $\Delta y(T_r, x_{N_2})$ \\
\hline

C$_5$H$_{12}$
& $0.091$
& $\phi_1^{(y)}$
& $\displaystyle
\Delta y
=
-
x_{N_2}
\underbrace{
\left(
a_1+e^{T_r}(a_2x_{N_2}+a_3)
\right)
}_{\text{(A) exponential nonlinear form}}
$ \\

C$_6$H$_{14}$
& $0.066$
& $\phi_2^{(y)}$
& $\displaystyle
\Delta y
=
\underbrace{
\left(
a_1-x_{N_2}-e^{a_2T_rx_{N_2}}
\right)
}_{\text{(B) linear--exponential form}}
$ \\

C$_7$H$_{16}$
& $0.072$
& $\phi_3^{(y)}$
& $\displaystyle
\Delta y
=
\underbrace{
\left(a_1T_r+a_2\right)
\left(x_{N_2}+a_3T_r^2\right)
}_{\text{(C) coupled polynomial form}}
$ \\

C$_9$H$_{20}$
& $0.029$
& $\phi_4^{(y)}$
& $\displaystyle
\Delta y
=
\underbrace{
\left(T_r+a_1\right)
\left(
\dfrac{x_{N_2}}{T_r-a_2}+a_3
\right)
}_{\text{(C) coupled polynomial form}}
$ \\

C$_{10}$H$_{22}$
& $0.014$
& $\phi_5^{(y)}$
& $\displaystyle
\Delta y
=
\left(
a_1
+
a_2
\hspace{-18pt}
\underbrace{
\left(
x_{N_2}-e^{x_{N_2}}
\right)
}_{\text{(B) linear--exponential form}}
\hspace{-18pt}
e^{a_3T_r}
\right)
\underbrace{
\left(
a_4+a_5x_{N_2}e^{T_r}
\right)
}_{\text{(A) exponential nonlinear form}}
+a_6
$ \\

C$_{12}$H$_{26}$
& $0.073$
& $\phi_6^{(y)}$
& $\displaystyle
\Delta y
=
a_1x_{N_2}
\underbrace{
\left(
e^{T_r}
\right)^{a_2^{x_{N_2}}}
}_{\text{(A) exponential nonlinear form}}
$ \\

\hline
\end{tabular}
\end{table}

As a result of the system-specific symbolic regression, distinct symbolic expressions are obtained for the pressure and vapor-composition correction terms of each hydrocarbon--nitrogen system, as summarized in Tables~\ref{tab:system_specific_normalized_dp} and \ref{tab:system_specific_normalized_dy}. Although the discovered expressions differ in both analytical form and complexity, several recurring symbolic forms consistently emerge across multiple hydrocarbon systems, including linear--exponential, coupled polynomial, exponential, and inverse-temperature forms. This observation suggests that the underlying correction mechanisms are not entirely system specific but instead share common symbolic structures with different numerical parameterizations. Motivated by this observation, the proposed multilevel symbolic regression extracts these recurring symbolic forms as shared basis functions while parameterizing only their coefficients as functions of the carbon number, thereby preserving analytical interpretability while enabling a unified symbolic correction model across multiple hydrocarbon--nitrogen systems.

\begin{figure}[htp!]
\centering

\begin{subfigure}[b]{1.0\linewidth}
    \centering
    \includegraphics[width=\linewidth]{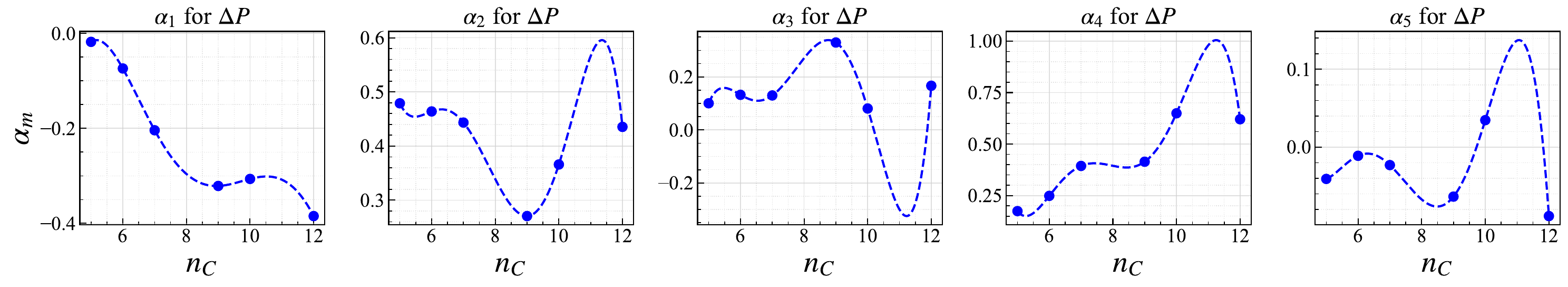}
    \caption{Polynomial interpolation of the coefficient functions $\alpha_m$ for $\Delta P$.}
\end{subfigure}
\begin{subfigure}[b]{1.0\linewidth}
    \centering
    \includegraphics[width=\linewidth]{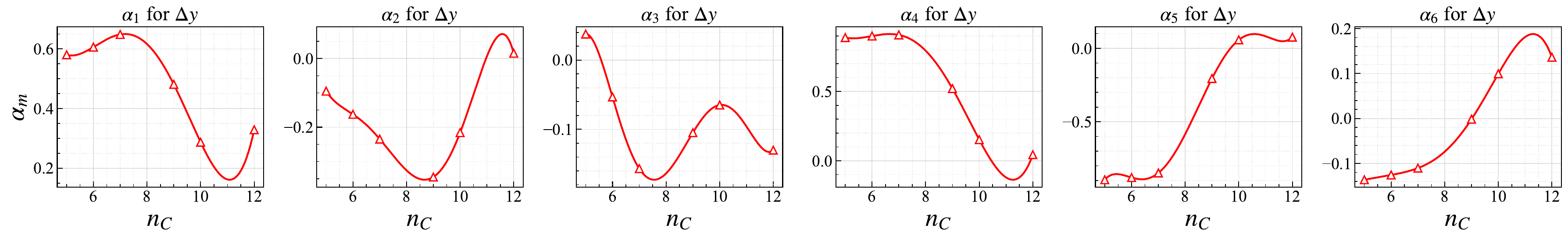}
    \caption{Polynomial interpolation of the coefficient functions $\alpha_m$ for $\Delta y$.}
\end{subfigure}
\caption{Polynomial interpolation of the learned systemwise coefficient values used in the shared-basis multilevel model. Symbols denote the coefficient values extracted from the inverse model, and solid curves denote the corresponding polynomial interpolants as functions of carbon number $n_C$.}
\label{fig:alpha_polynomial_interpolants}
\end{figure}

To enable prediction for different hydrocarbon systems using a common symbolic basis, the system-specific coefficients are parameterized as continuous functions of the carbon number. The coefficient values identified for each binary system are first extracted from the multilevel symbolic regression model and subsequently approximated using polynomial interpolation, as illustrated in Fig.~\ref{fig:alpha_polynomial_interpolants}. Separate interpolants are constructed for the pressure correction coefficients, as shown in Fig.~\ref{fig:alpha_polynomial_interpolants}a, and for the vapor-composition correction coefficients, as shown in Fig.~\ref{fig:alpha_polynomial_interpolants}b. As a result, only the coefficient values vary systematically with the carbon number, whereas the shared symbolic basis functions remain unchanged. This formulation enables the proposed model to represent multiple hydrocarbon--nitrogen systems within a unified symbolic expression while preserving the interpretability of the discovered symbolic correction model.


\begin{figure}[htp!]
\centering
\includegraphics[width=0.95\linewidth]{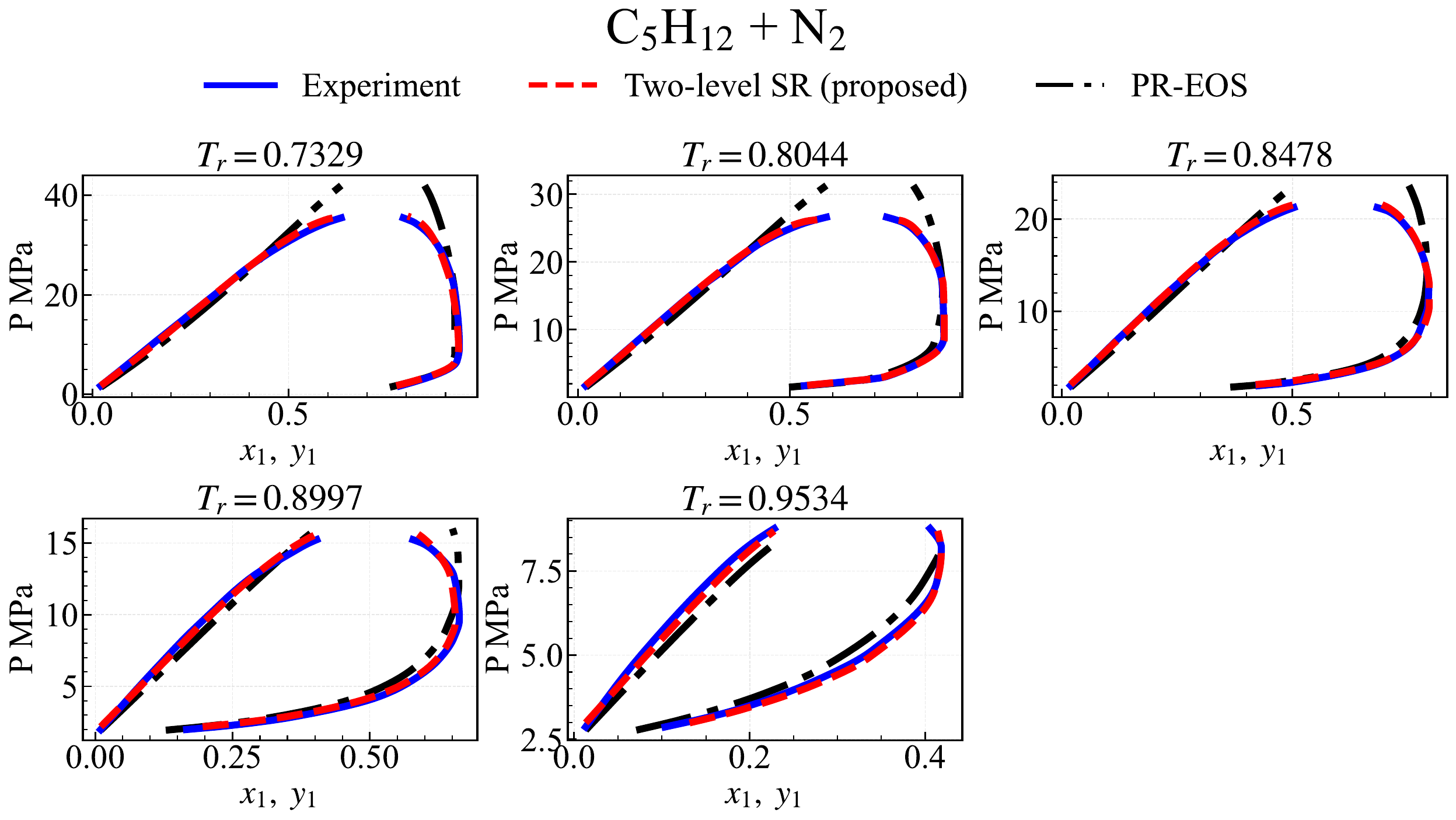}
\caption{Pressure-composition diagrams for the C$_5$H$_{12}$ + N$_2$ system at different reduced temperatures. The experimental VLE curves, the PR-EOS predictions, and the multilevel regression predictions are compared in each temperature panel.}
\label{fig:c5n2_vle_temp_panels}
\end{figure}

\begin{figure}[htp!]
\centering
\includegraphics[width=0.95\linewidth]{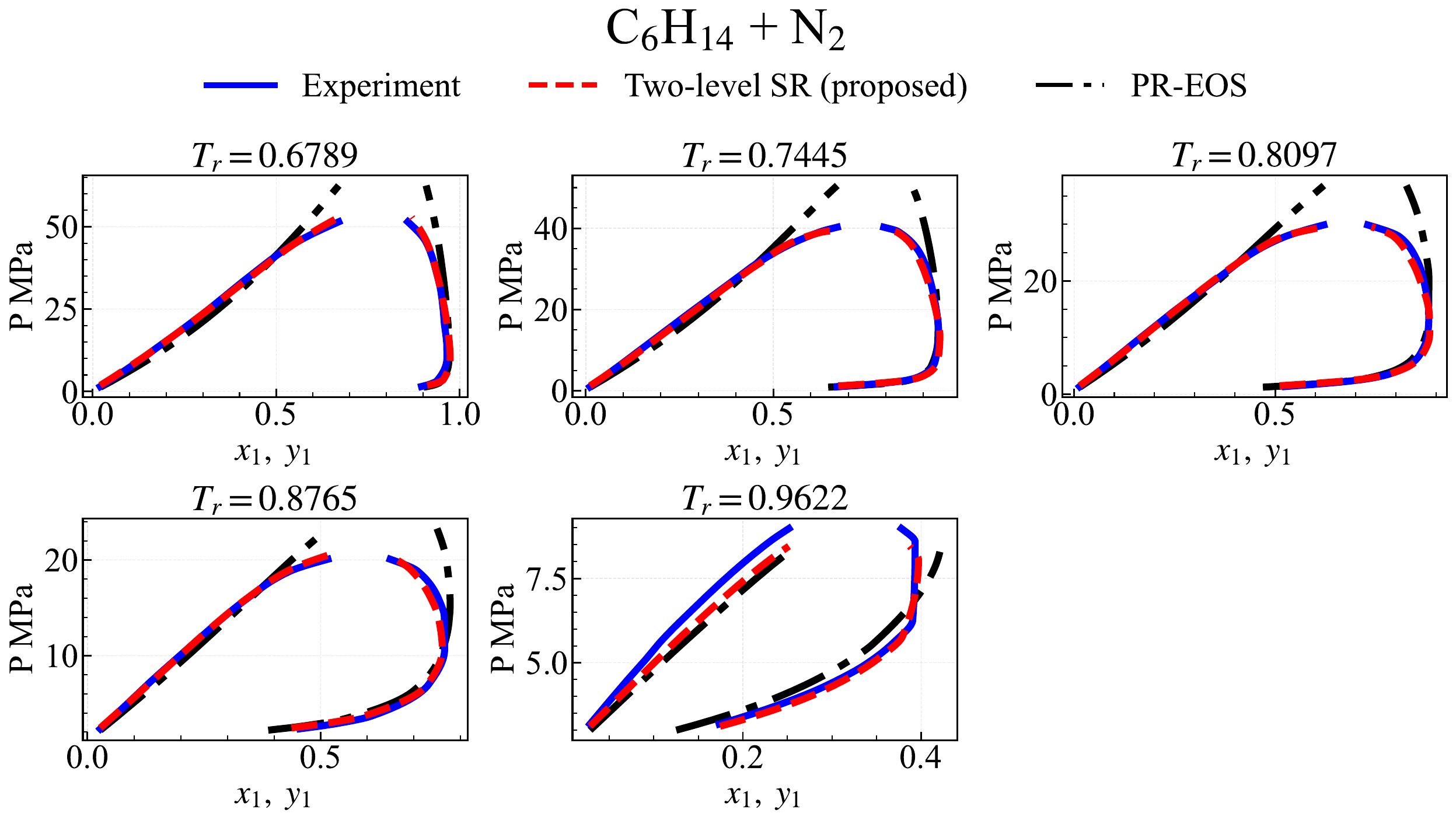}
\caption{Pressure-composition diagrams for the C$_6$H$_{14}$ + N$_2$ system at different reduced temperatures. The experimental VLE curves, the PR-EOS predictions, and the multilevel regression predictions are compared in each temperature panel.}
\label{fig:c6n2_vle_temp_panels}
\end{figure}

\begin{figure}[htp!]
\centering
\includegraphics[width=0.95\linewidth]{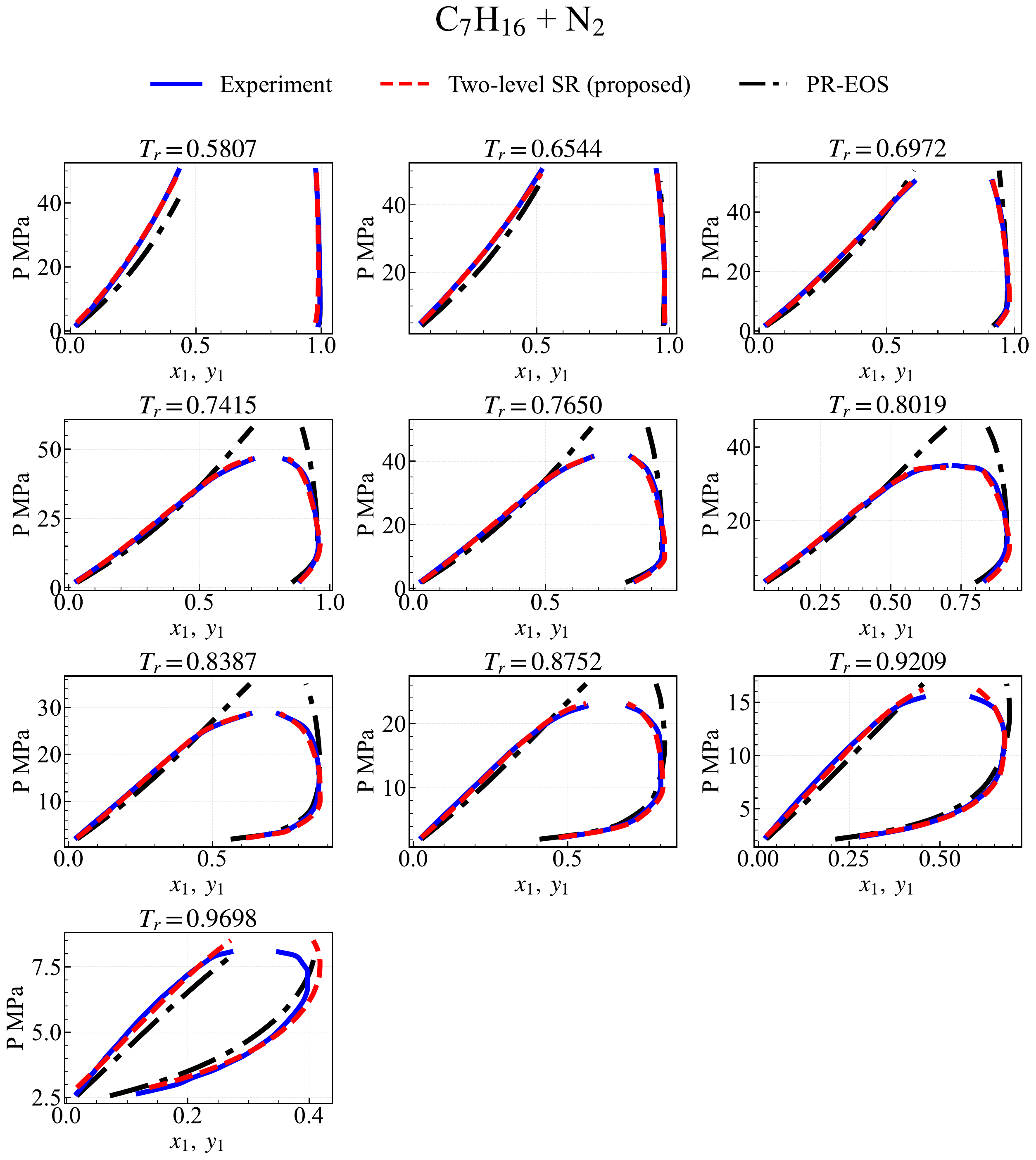}
\caption{Pressure-composition diagrams for the C$_7$H$_{16}$ + N$_2$ system at different reduced temperatures. The experimental VLE curves, the PR-EOS predictions, and the multilevel regression predictions are compared in each temperature panel.}
\label{fig:c7n2_vle_temp_panels}
\end{figure}

\begin{figure}[htp!]
\centering
\includegraphics[width=0.95\linewidth]{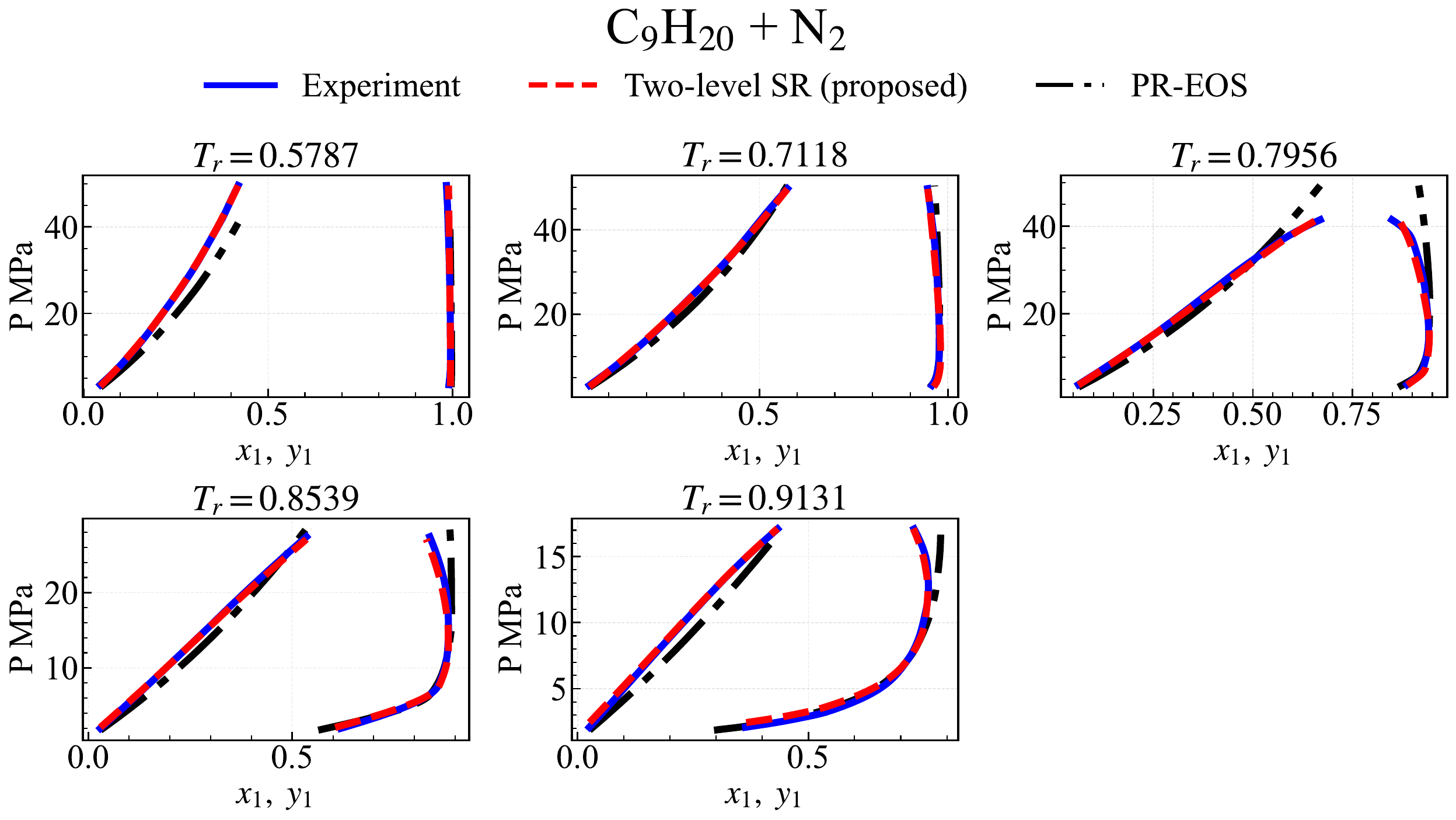}
\caption{Pressure-composition diagrams for the C$_9$H$_{20}$ + N$_2$ system at different reduced temperatures. The experimental VLE curves, the PR-EOS predictions, and the multilevel regression predictions are compared in each temperature panel.}
\label{fig:c9n2_vle_temp_panels}
\end{figure}

\begin{figure}[htp!]
\centering
\includegraphics[width=0.95\linewidth]{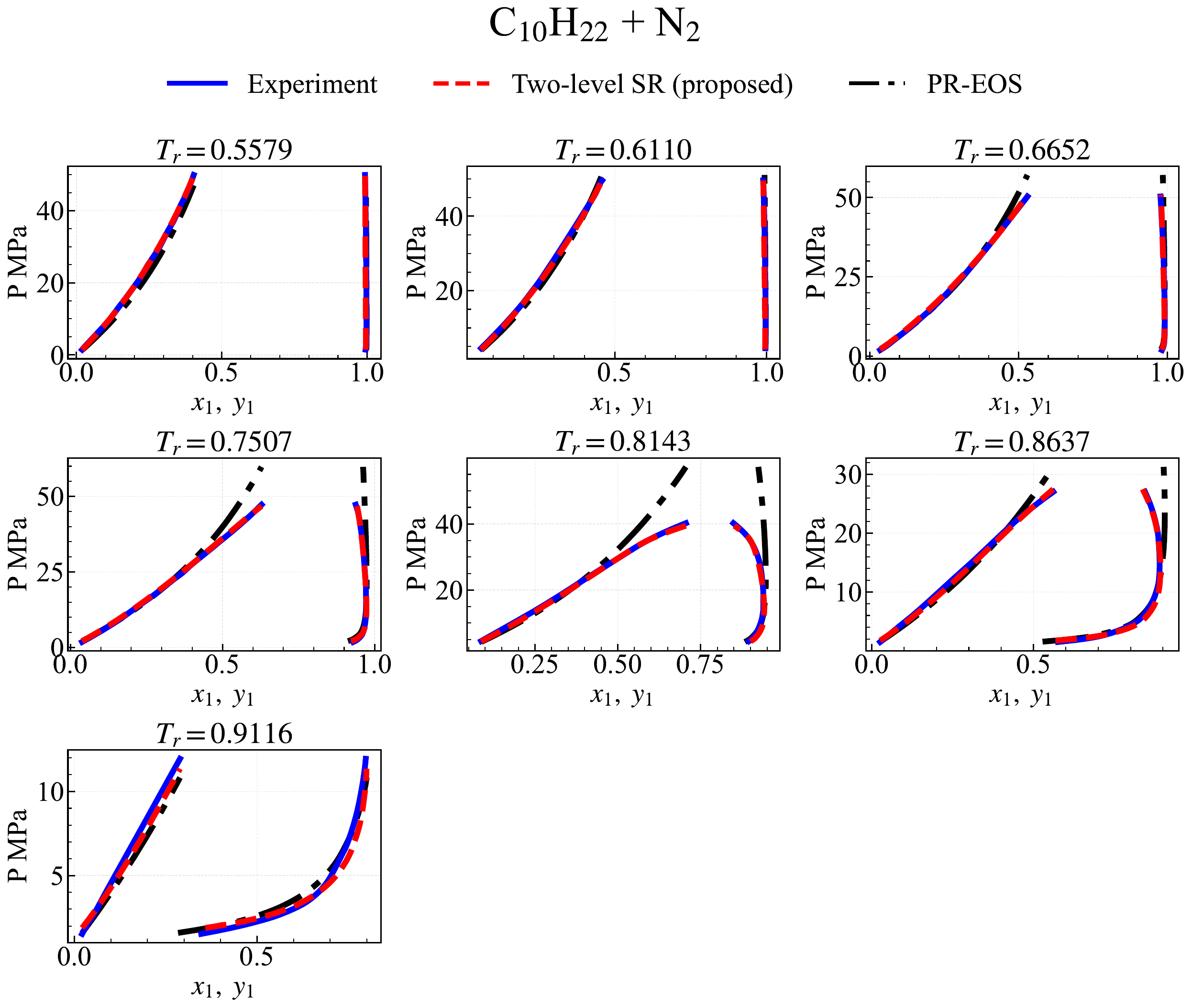}
\caption{Pressure-composition diagrams for the C$_{10}$H$_{22}$ + N$_2$ system at different reduced temperatures. The experimental VLE curves, the PR-EOS predictions, and the multilevel regression predictions are compared in each temperature panel.}
\label{fig:c10n2_vle_temp_panels}
\end{figure}

\begin{figure}[htp!]
\centering
\includegraphics[width=0.95\linewidth]{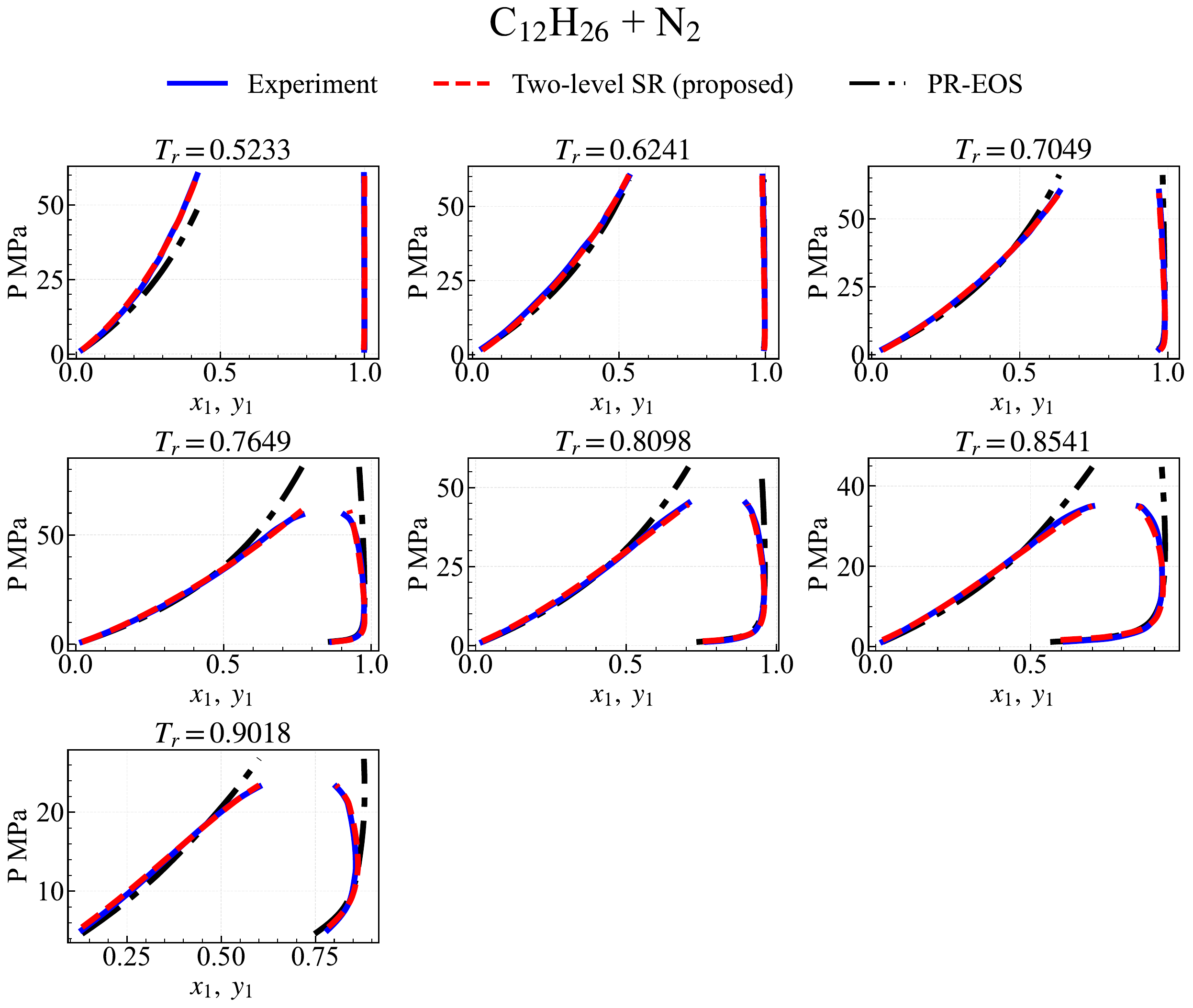}
\caption{Pressure-composition diagrams for the C$_{12}$H$_{26}$ + N$_2$ system at different reduced temperatures. The experimental VLE curves, the PR-EOS predictions, and the multilevel regression predictions are compared in each temperature panel.}
\label{fig:c12n2_vle_temp_panels}
\end{figure}

To evaluate the predictive performance of the proposed multilevel symbolic regression model, the corrected PR-EOS predictions are compared with both the experimental VLE measurements and the original PR-EOS for all six hydrocarbon--nitrogen systems over a wide range of reduced temperatures, as shown in Figs.~\ref{fig:c5n2_vle_temp_panels}--\ref{fig:c12n2_vle_temp_panels}. Overall, the proposed symbolic correction consistently improves agreement with the experimental pressure--composition curves while preserving the physically consistent trends predicted by the underlying PR-EOS. Across all hydrocarbon systems, the corrected model accurately reproduces the pressure variation over the entire composition range and substantially reduces the systematic deviations observed in the original PR-EOS, demonstrating that the learned symbolic corrections effectively capture the dominant modeling errors.

For the lighter hydrocarbon systems, including C$_5$H$_{12}$, C$_6$H$_{14}$, and C$_7$H$_{16}$, the proposed model accurately reproduces the experimental phase-equilibrium curves over all investigated reduced temperatures, as shown in Figs.~\ref{fig:c5n2_vle_temp_panels}--\ref{fig:c7n2_vle_temp_panels}. Although the original PR-EOS already provides reasonable agreement with the experimental data, noticeable discrepancies remain, particularly in regions exhibiting strong nonlinear pressure variation. The proposed symbolic correction consistently reduces these deviations while preserving the smooth thermodynamic trends of the original equation of state, resulting in excellent agreement with the measured VLE curves throughout the entire composition range.

The improvement becomes even more evident for the heavier hydrocarbon systems, including C$_9$H$_{20}$, C$_{10}$H$_{22}$, and C$_{12}$H$_{26}$, as illustrated in Figs.~\ref{fig:c9n2_vle_temp_panels}--\ref{fig:c12n2_vle_temp_panels}. For these systems, the original PR-EOS exhibits larger systematic deviations from the experimental measurements, particularly at elevated pressures. By incorporating the learned symbolic correction, the proposed model substantially improves the agreement with the experimental pressure--composition curves while maintaining smooth and physically consistent predictions across different reduced temperatures. These results demonstrate that the proposed multilevel symbolic regression successfully captures systematic PR-EOS errors over a broad range of hydrocarbon chain lengths using a unified symbolic correction model.

\begin{table}[htp!]
\footnotesize
\centering
\caption{Accuracy comparison for pressure prediction for the hydrocarbon--nitrogen VLE dataset. Train and test denote the common pointwise 80/20 split used for evaluation. Reported metrics are the mean squared error (MSE), coefficient of determination ($R^2$), maximum absolute error, and mean absolute error. The best value in each column is shown in bold.}
\label{tab:vle_multilevel_metrics_pressure}
\begin{tabular}{llcccc}
\hline
Model & Split & MSE & $R^2$ & $\max |P-\hat P|$ & $\mathrm{mean}\,|P-\hat P|$ \\
\hline
\multirow{2}{*}{Two-level SR (proposed)}
& Train & $\mathbf{1.1085\times10^{-1}}$ & $\mathbf{0.9995}$ & $\mathbf{2.0615\times10^{0}}$ & $\mathbf{2.5215\times10^{-1}}$ \\
& Test  & $\mathbf{1.2279\times10^{-1}}$ & $\mathbf{0.9994}$ & $\mathbf{8.1785\times10^{-1}}$ & $\mathbf{2.8475\times10^{-1}}$ \\
\hline
\multirow{2}{*}{SR}
& Train & $1.4301\times10^{0}$ & $0.9941$ & $5.9095\times10^{0}$ & $7.5660\times10^{-1}$ \\
& Test  & $7.3063\times10^{-1}$ & $0.9963$ & $2.8557\times10^{0}$ & $6.1370\times10^{-1}$ \\
\hline
\multirow{2}{*}{PR-EOS}
& Train & $1.2277\times10^{1}$ & $0.9489$ & $2.1876\times10^{1}$ & $2.0157\times10^{0}$ \\
& Test  & $9.3985\times10^{0}$ & $0.9528$ & $1.2690\times10^{1}$ & $1.8271\times10^{0}$ \\
\hline
\end{tabular}
\end{table}

To quantitatively evaluate the effectiveness of the proposed multilevel symbolic regression, Table~\ref{tab:vle_multilevel_metrics_pressure} compares its pressure prediction accuracy with those of the conventional symbolic regression model and the original PR-EOS using a common 80/20 train--test split. The proposed two-level symbolic regression consistently achieves the lowest prediction errors and the highest coefficient of determination on both the training and test sets. On the test set, the mean squared error is reduced from $7.31\times10^{-1}$ for the conventional symbolic regression to $1.23\times10^{-1}$, while the mean absolute error decreases from $6.14\times10^{-1}$ to $2.85\times10^{-1}$. Compared with the original PR-EOS, the improvement is even more substantial, reducing the mean squared error by nearly two orders of magnitude. These results demonstrate that introducing shared symbolic basis functions together with carbon-number-dependent coefficient functions significantly improves prediction accuracy without sacrificing generalization performance.

\begin{table}[htp!]
\footnotesize
\centering
\caption{Accuracy comparison for vapor-phase nitrogen composition prediction for the hydrocarbon--nitrogen VLE dataset. Train and test denote the common pointwise 80/20 split used for evaluation. Reported metrics are the mean squared error (MSE), coefficient of determination ($R^2$), maximum absolute error, and mean absolute error. The best value in each column is shown in bold.}
\label{tab:vle_multilevel_metrics_y}
\begin{tabular}{llcccc}
\hline
Model & Split & MSE & $R^2$ & $\max |y_{N_2}-\hat y_{N_2}|$ & $\mathrm{mean}\,|y_{N_2}-\hat y_{N_2}|$ \\
\hline
\multirow{2}{*}{Two-level SR (proposed)}
& Train & $\mathbf{5.6785\times10^{-5}}$ & $\mathbf{0.9987}$ & $\mathbf{5.7109\times10^{-2}}$ & $\mathbf{4.7903\times10^{-3}}$ \\
& Test  & $\mathbf{3.7679\times10^{-5}}$ & $\mathbf{0.9986}$ & $\mathbf{1.1891\times10^{-2}}$  & $\mathbf{5.0269\times10^{-3}}$ \\
\hline
\multirow{2}{*}{SR}
& Train & $1.2469\times10^{-4}$ & $0.9971$ & $6.6379\times10^{-2}$ & $7.4529\times10^{-3}$ \\
& Test  & $4.8733\times10^{-5}$ & $0.9982$ & $1.5917\times10^{-2}$& $5.0378\times10^{-3}$ \\
\hline
\multirow{2}{*}{PR-EOS}
& Train & $1.0735\times10^{-3}$ & $0.9750$ & $1.1880\times10^{-1}$ & $2.1437\times10^{-2}$ \\
& Test  & $6.4073\times10^{-4}$ & $0.9762$ & $5.6650\times10^{-2}$ & $1.7970\times10^{-2}$ \\
\hline
\end{tabular}
\end{table}

To verify that the proposed multilevel symbolic regression accurately predicts the equilibrium composition in addition to pressure, Table~\ref{tab:vle_multilevel_metrics_y} summarizes the prediction accuracy for the vapor-phase nitrogen mole fraction. Similar to the pressure prediction results, the proposed model consistently achieves the lowest MSE and the highest $R^2$ on both the training and test sets. Compared with the original PR-EOS, the prediction errors are substantially reduced, while modest but consistent improvements over the conventional symbolic regression further demonstrate the benefit of the proposed two-level symbolic formulation.

\begin{figure}[htp!]
\centering

\begin{subfigure}[b]{0.44\linewidth}
    \centering
    \includegraphics[width=\linewidth]{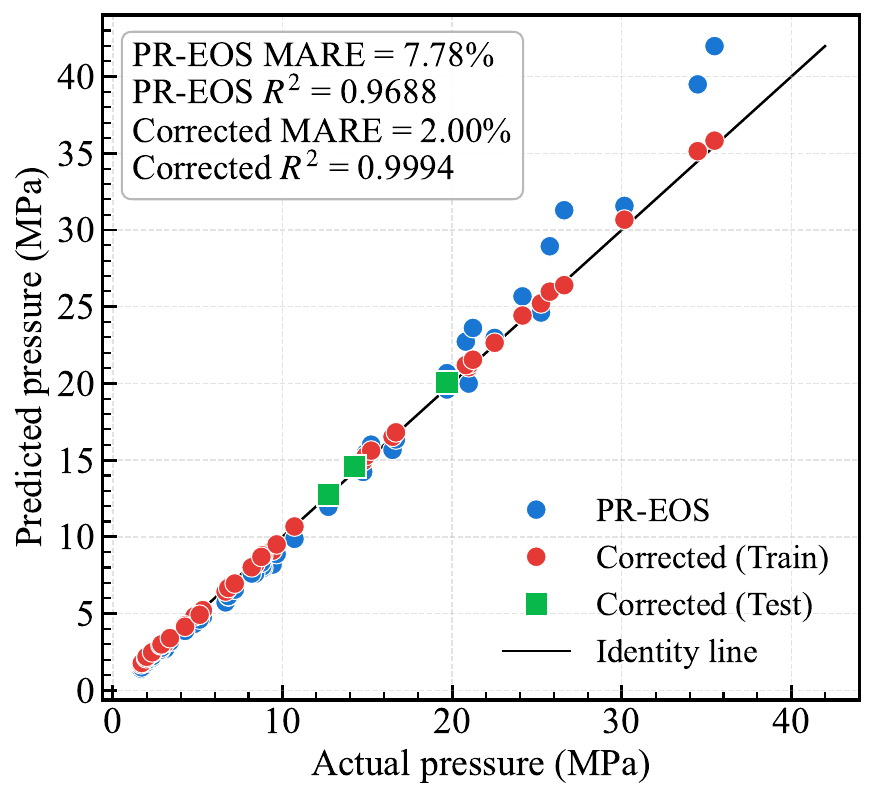}
    \caption{C$_5$H$_{12}$ + N$_2$}
\end{subfigure}
\begin{subfigure}[b]{0.44\linewidth}
    \centering
    \includegraphics[width=\linewidth]{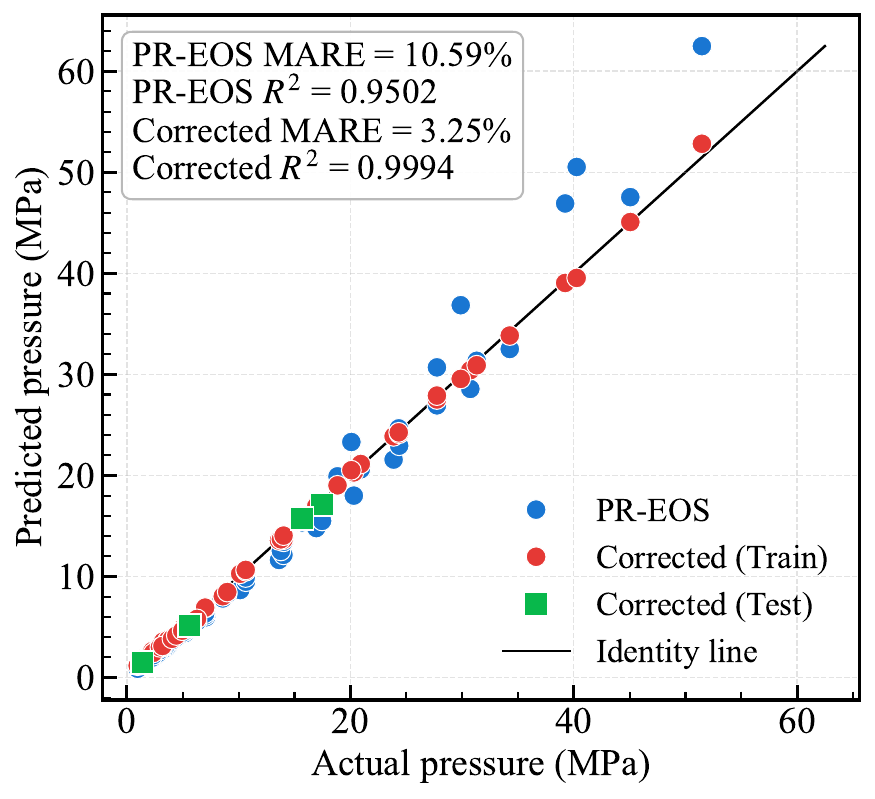}
    \caption{C$_6$H$_{14}$ + N$_2$}
\end{subfigure}
\\
\begin{subfigure}[b]{0.44\linewidth}
    \centering
    \includegraphics[width=\linewidth]{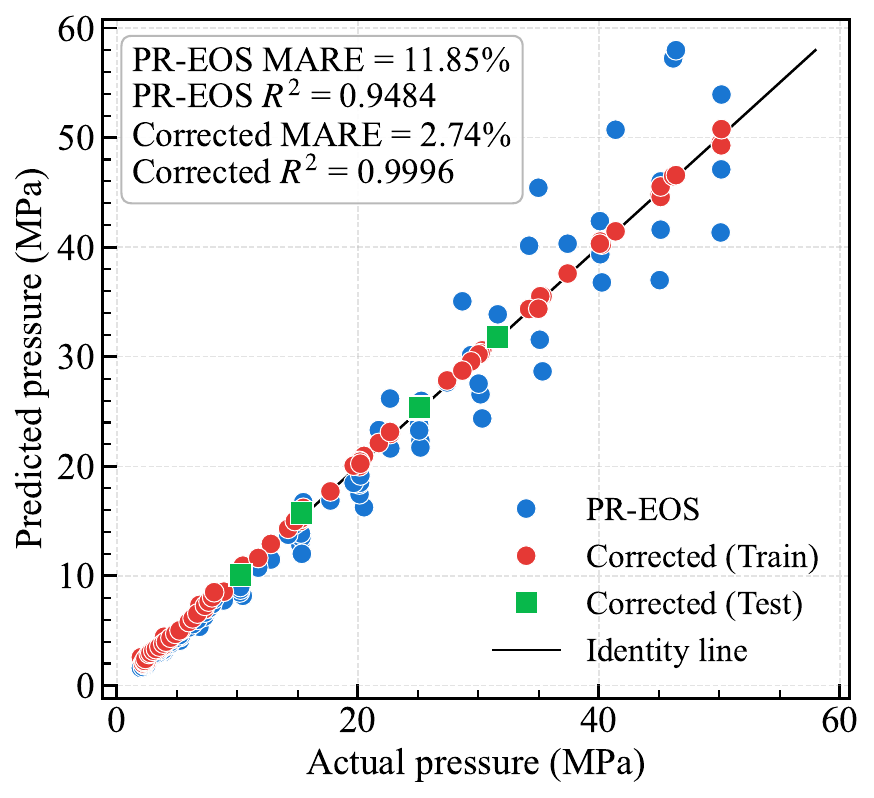}
    \caption{C$_7$H$_{16}$ + N$_2$}
\end{subfigure}
\begin{subfigure}[b]{0.44\linewidth}
    \centering
    \includegraphics[width=\linewidth]{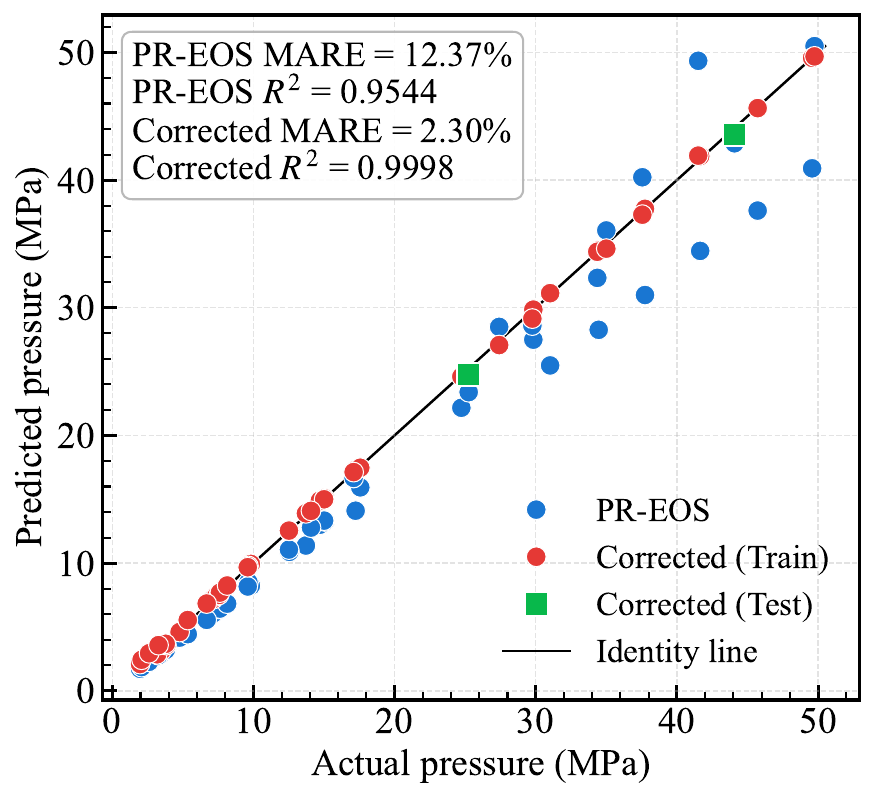}
    \caption{C$_9$H$_{20}$ + N$_2$}
\end{subfigure}
\\
\begin{subfigure}[b]{0.44\linewidth}
    \centering
    \includegraphics[width=\linewidth]{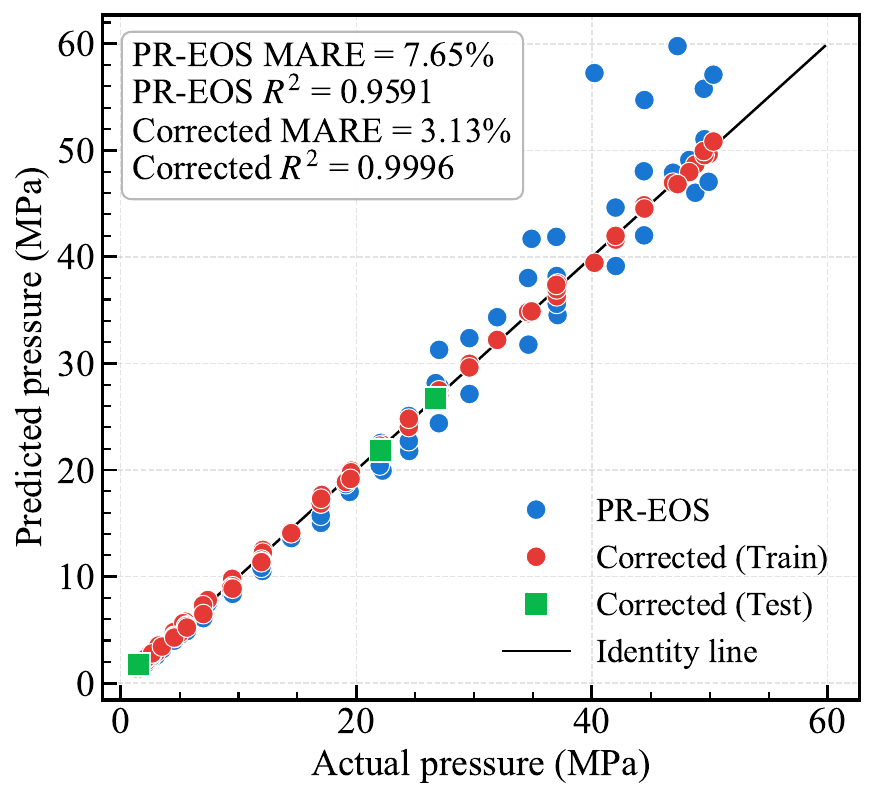}
    \caption{C$_{10}$H$_{22}$ + N$_2$}
\end{subfigure}
\begin{subfigure}[b]{0.44\linewidth}
    \centering
    \includegraphics[width=\linewidth]{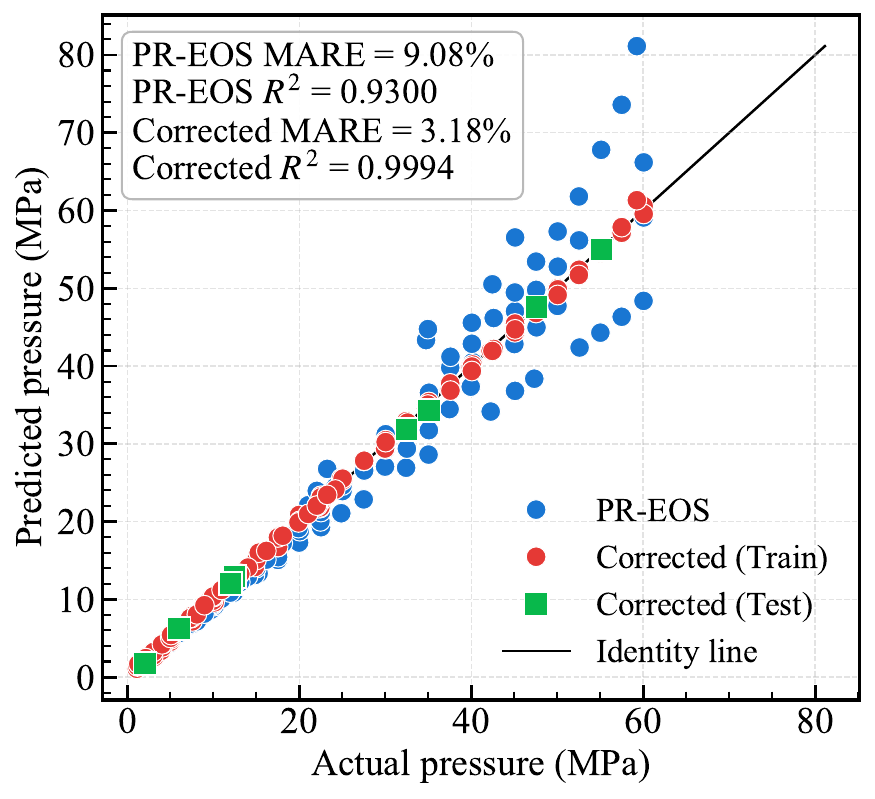}
    \caption{C$_{12}$H$_{26}$ + N$_2$}
\end{subfigure}
\caption{
Actual-versus-predicted pressure plots for the six hydrocarbon--nitrogen systems. Black circles denote the original PR-EOS predictions, red circles denote corrected training samples, green squares denote corrected test samples, and the black line represents the identity line.
}
\label{fig:pressure_parity_lookup_grid}
\end{figure}

To further evaluate the pressure prediction accuracy, the actual and predicted pressures are compared for all six hydrocarbon--nitrogen systems in Fig.~\ref{fig:pressure_parity_lookup_grid}. Compared with the original PR-EOS predictions, the proposed symbolic correction substantially reduces the prediction error, with both the training and test samples becoming much more closely aligned with the identity line. The improvement is consistently observed across all six hydrocarbon systems, indicating that the proposed correction effectively captures the systematic pressure deviations of the PR-EOS while maintaining strong predictive performance on unseen test samples.

\begin{figure}[htp!]
\centering
\begin{subfigure}[b]{0.44\linewidth}
    \centering
    \includegraphics[width=\linewidth]{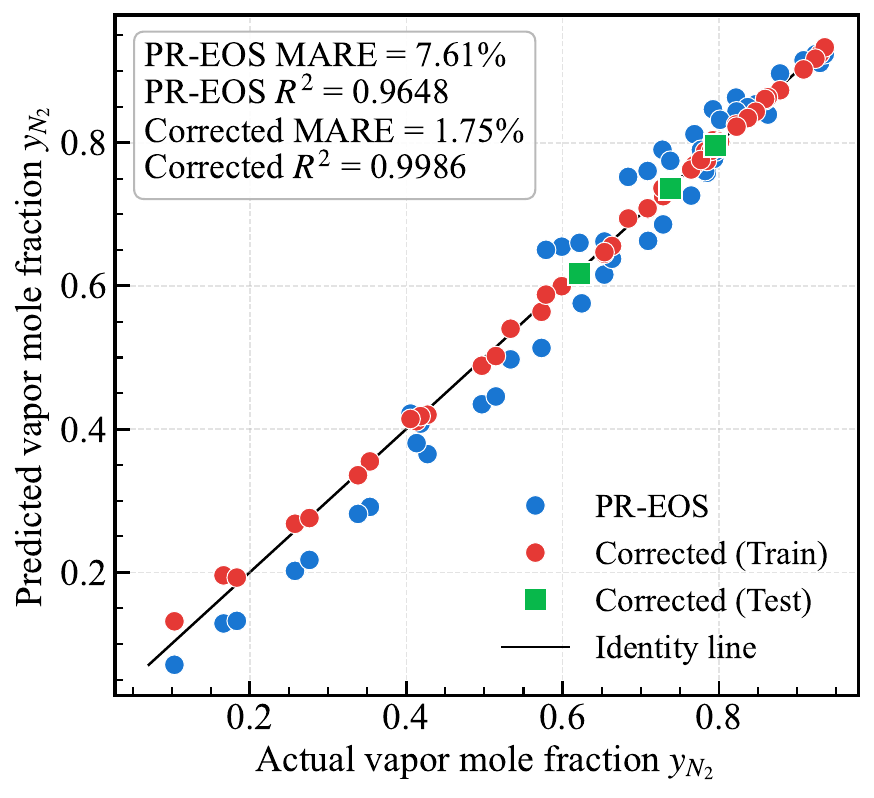}
    \caption{C$_5$H$_{12}$ + N$_2$}
\end{subfigure}
\begin{subfigure}[b]{0.44\linewidth}
    \centering
    \includegraphics[width=\linewidth]{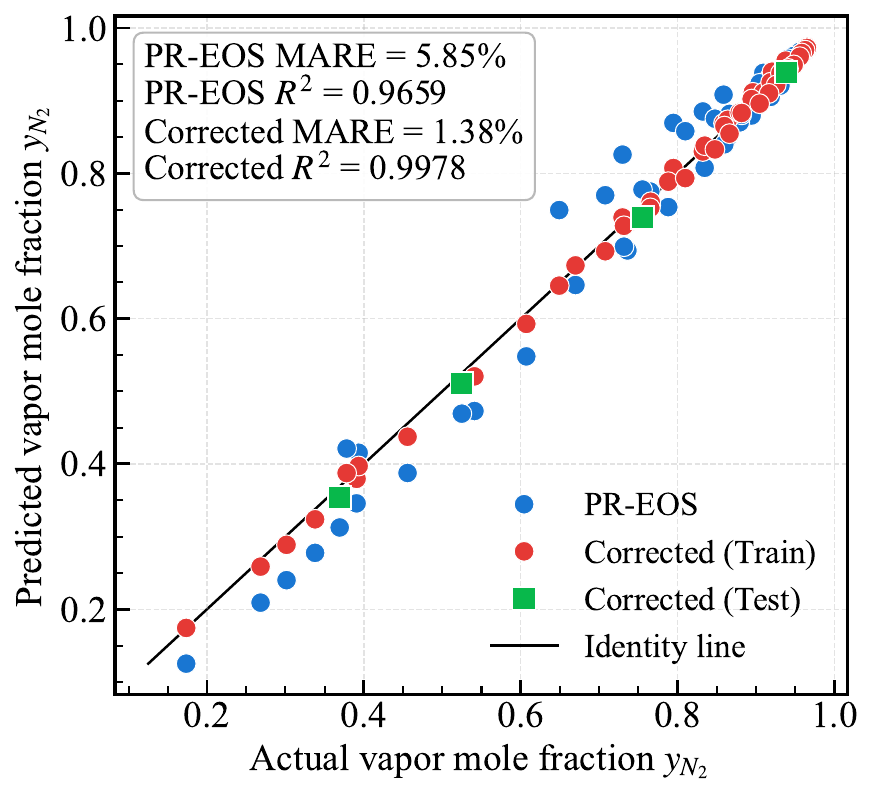}
    \caption{C$_6$H$_{14}$ + N$_2$}
\end{subfigure}
\\
\begin{subfigure}[b]{0.44\linewidth}
    \centering
    \includegraphics[width=\linewidth]{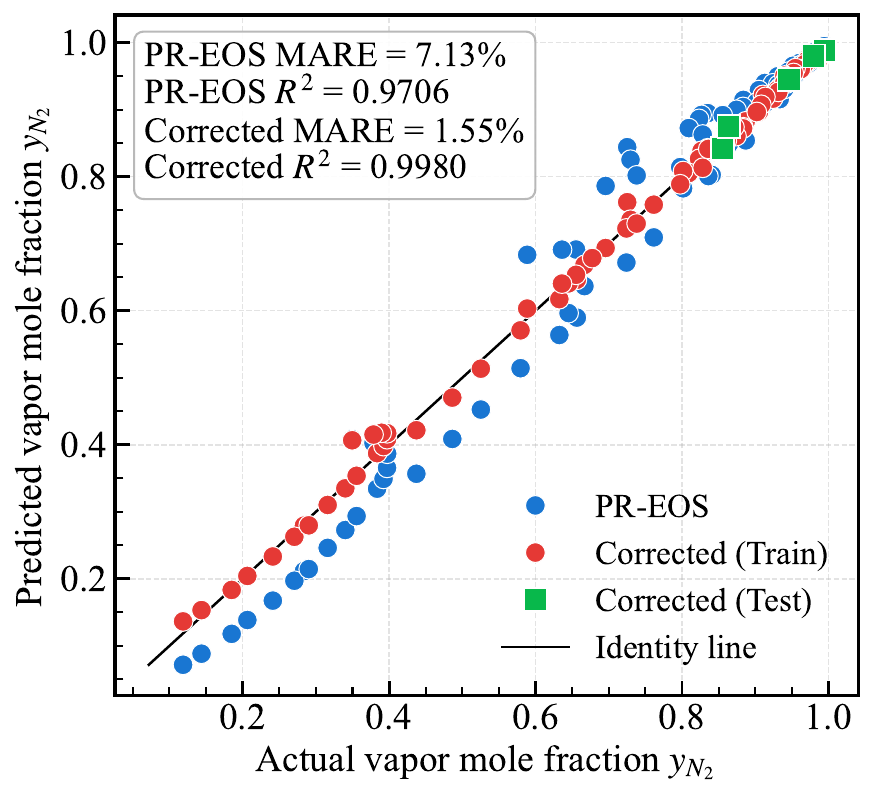}
    \caption{C$_7$H$_{16}$ + N$_2$}
\end{subfigure}
\begin{subfigure}[b]{0.44\linewidth}
    \centering
    \includegraphics[width=\linewidth]{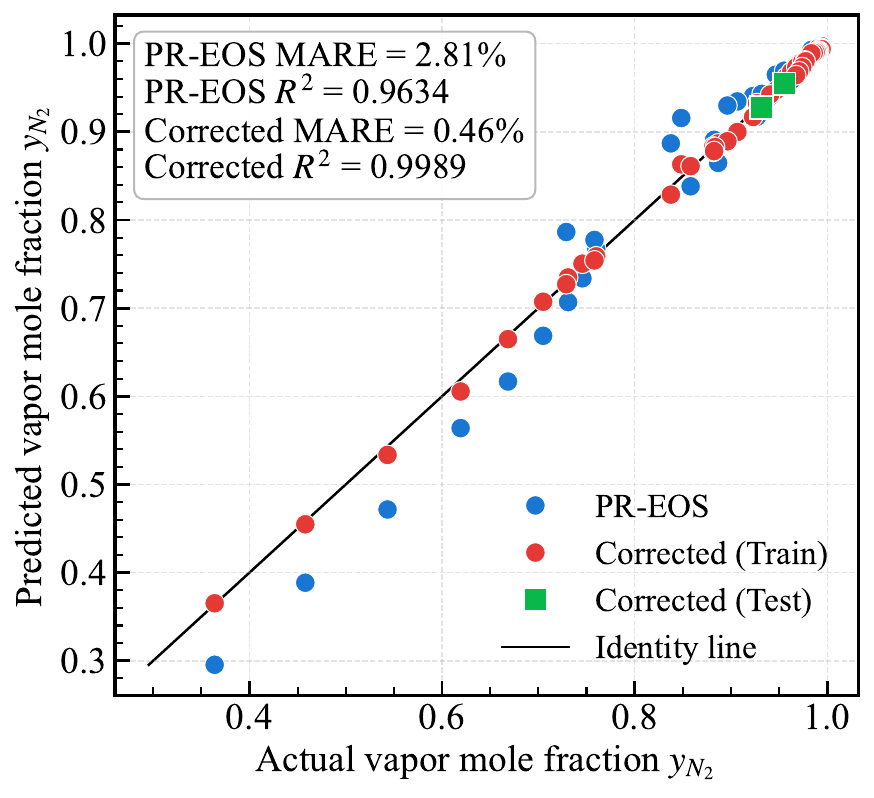}
    \caption{C$_9$H$_{20}$ + N$_2$}
\end{subfigure}
\\
\begin{subfigure}[b]{0.44\linewidth}
    \centering
    \includegraphics[width=\linewidth]{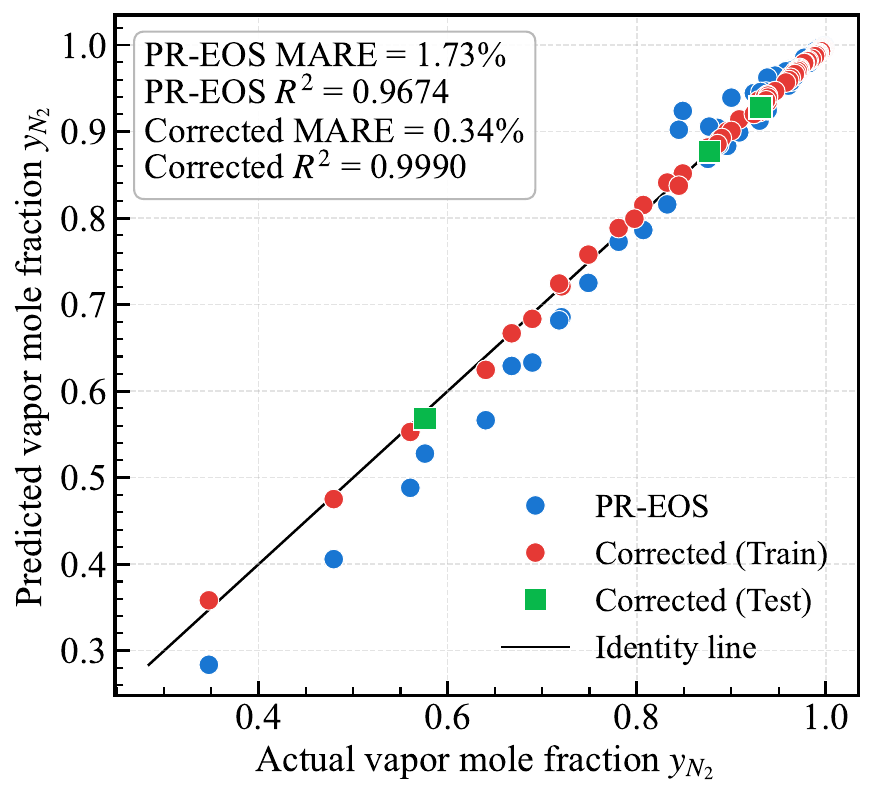}
    \caption{C$_{10}$H$_{22}$ + N$_2$}
\end{subfigure}
\begin{subfigure}[b]{0.44\linewidth}
    \centering
    \includegraphics[width=\linewidth]{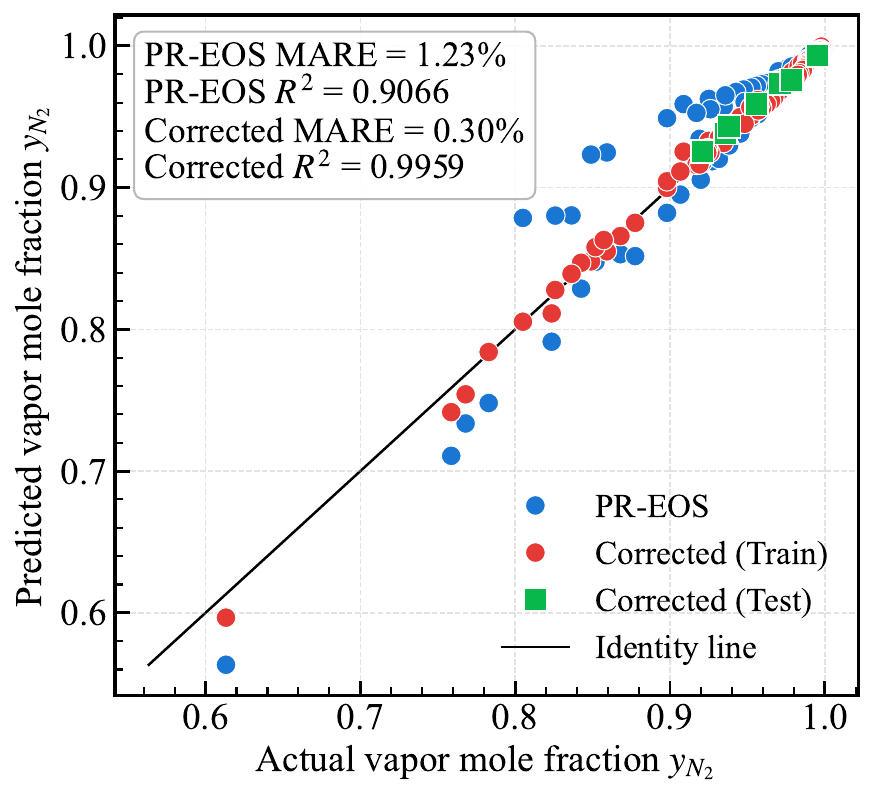}
    \caption{C$_{12}$H$_{26}$ + N$_2$}
\end{subfigure}
\caption{Vapor composition parity plots for the lookup coefficient model across the six hydrocarbon--nitrogen systems. black circles denote the original PR-EOS predictions, red circles denote corrected training samples, magenta squares denote corrected test samples, and the black line denotes the identity line.}
\label{fig:y_parity_lookup_grid}
\end{figure}

Figure~\ref{fig:y_parity_lookup_grid} presents the vapor composition parity plots for all six hydrocarbon--nitrogen systems. Compared with the original PR-EOS predictions, the proposed lookup coefficient model significantly improves the prediction accuracy, with both the training and test samples closely distributed along the identity line. The consistent agreement across all systems demonstrates that the proposed correction model accurately captures the systematic vapor composition deviations of the PR-EOS while maintaining good generalization performance.

\begin{figure}[htp!]
\centering

\includegraphics[width=0.95\linewidth]{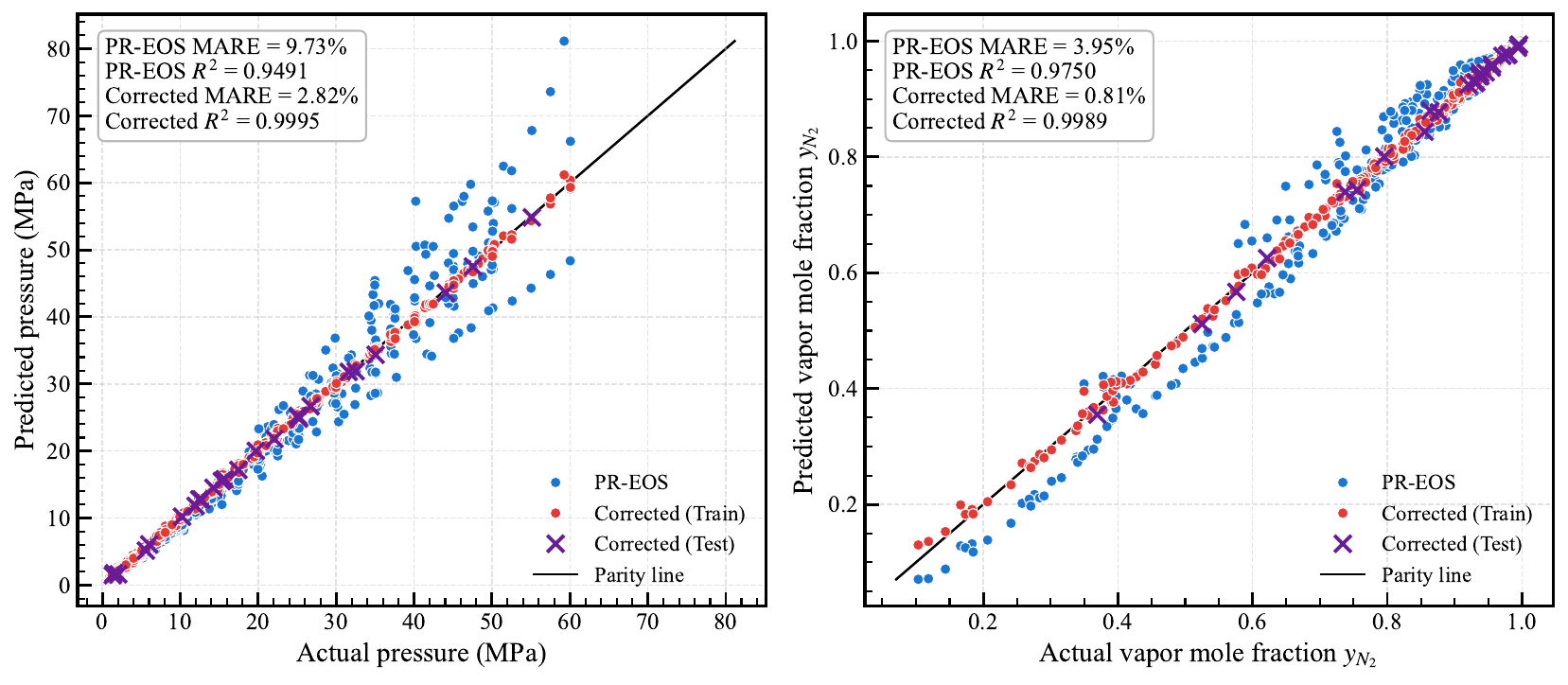}
\caption{
Actual-versus-predicted plots for pressure (left) and vapor mole fraction (right) obtained by combining all six hydrocarbon--nitrogen systems. Black circles denote the original PR-EOS predictions, red crosses denote corrected training samples, green squares denote corrected test samples, and the solid black line represents the identity line.
}
\label{fig:actual_predicted_whole}
\end{figure}

To provide an overall assessment of the proposed symbolic correction model, Fig.~\ref{fig:actual_predicted_whole} combines the prediction results from all six hydrocarbon--nitrogen systems into unified actual-versus-predicted plots for pressure and vapor mole fraction. Compared with the original PR-EOS predictions, the corrected training and test samples exhibit substantially improved agreement with the identity line for both quantities. In particular, the pressure predictions are significantly improved over the entire pressure range, while the vapor-phase composition predictions remain highly accurate across a broad range of mixture compositions. The close agreement observed for both the training and test samples demonstrates that the proposed multilevel symbolic regression consistently improves predictive accuracy while maintaining robust generalization across different hydrocarbon systems.

\begin{figure}[htp!]
\centering
\includegraphics[width=0.95\linewidth]{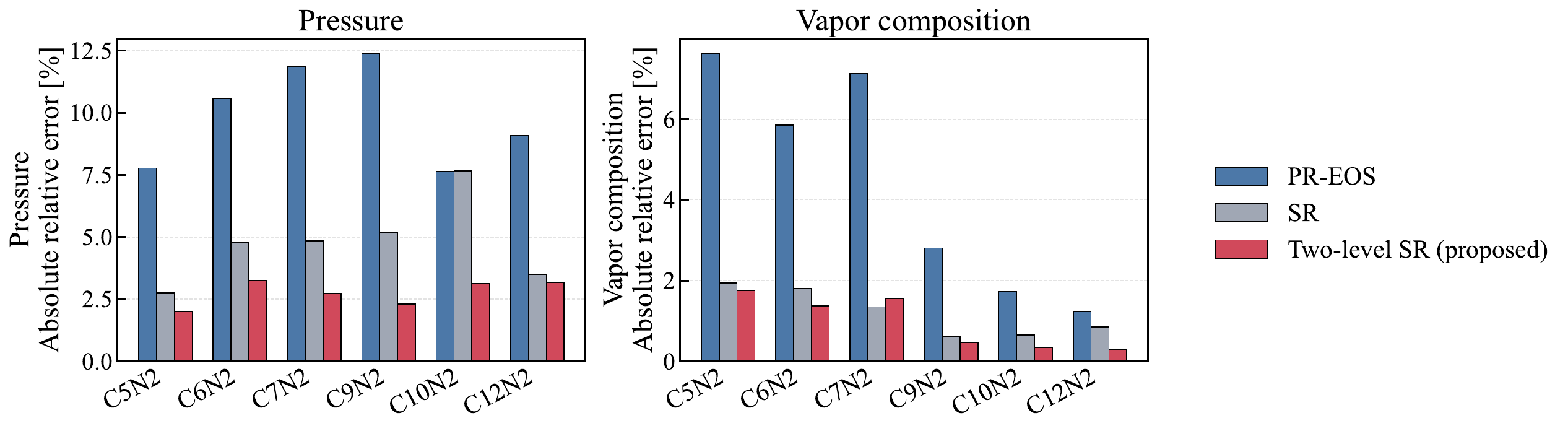}
\includegraphics[width=0.95\linewidth]{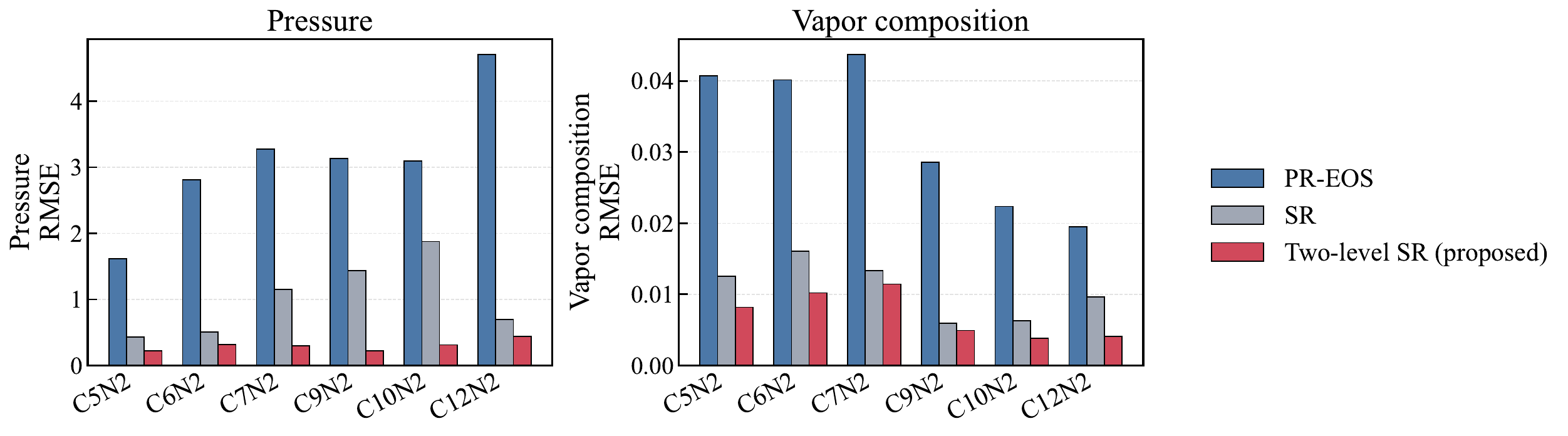}
\includegraphics[width=0.95\linewidth]{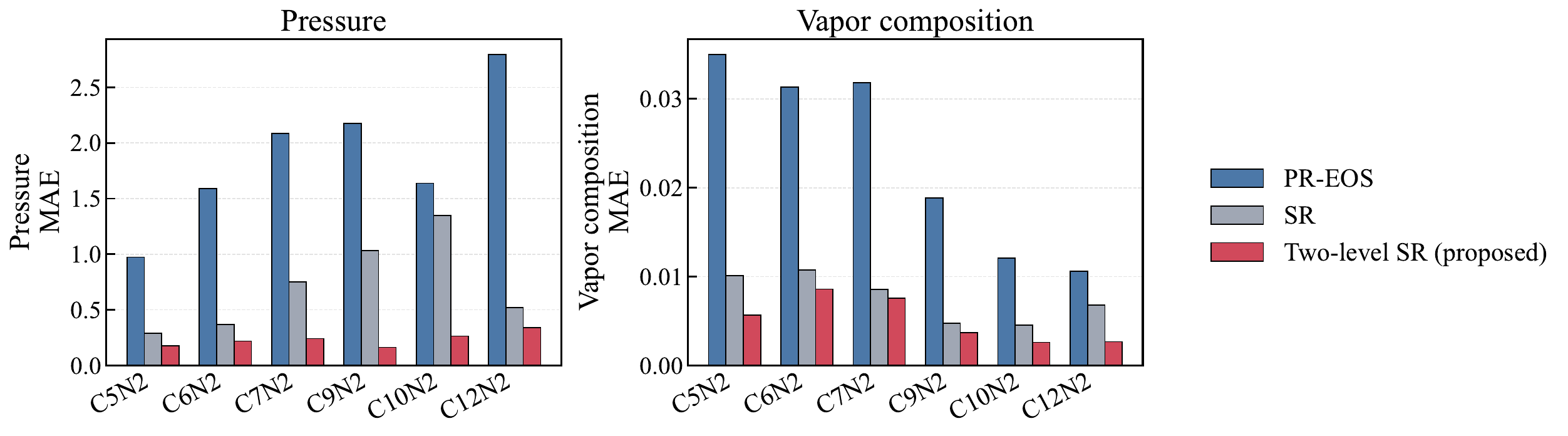}
\caption{
Systemwise comparison of prediction errors for pressure and vapor composition across the six hydrocarbon--nitrogen mixtures. 
The top panel shows the absolute relative error, the middle panel shows the root-mean-square error (RMSE), and the bottom panel shows the mean absolute error (MAE). 
In each case, the original PR-EOS predictions, the pooled symbolic regression model, and the proposed multilevel regression model are compared for every system.
}
\label{fig:error_rmse_mae_comparison}
\end{figure}

To assess the effectiveness of the proposed multilevel symbolic regression, Fig.~\ref{fig:error_rmse_mae_comparison} compares the prediction errors of the original PR-EOS, the conventional symbolic regression model, and the proposed two-level symbolic regression model across all six hydrocarbon--nitrogen systems. The absolute relative error, RMSE, and MAE are reported for both pressure and vapor composition.
Compared with the original PR-EOS, both symbolic regression models substantially reduce the prediction errors. More importantly, the proposed two-level symbolic regression consistently outperforms the conventional symbolic regression across nearly all hydrocarbon systems and error metrics. These results demonstrate that separating system-specific symbolic discovery from carbon-number-dependent coefficient learning provides a more accurate symbolic representation than learning a single symbolic expression from the pooled dataset.

\begin{figure}[htp!]
\centering
\begin{subfigure}[b]{0.9\linewidth}
    \includegraphics[width=\linewidth]{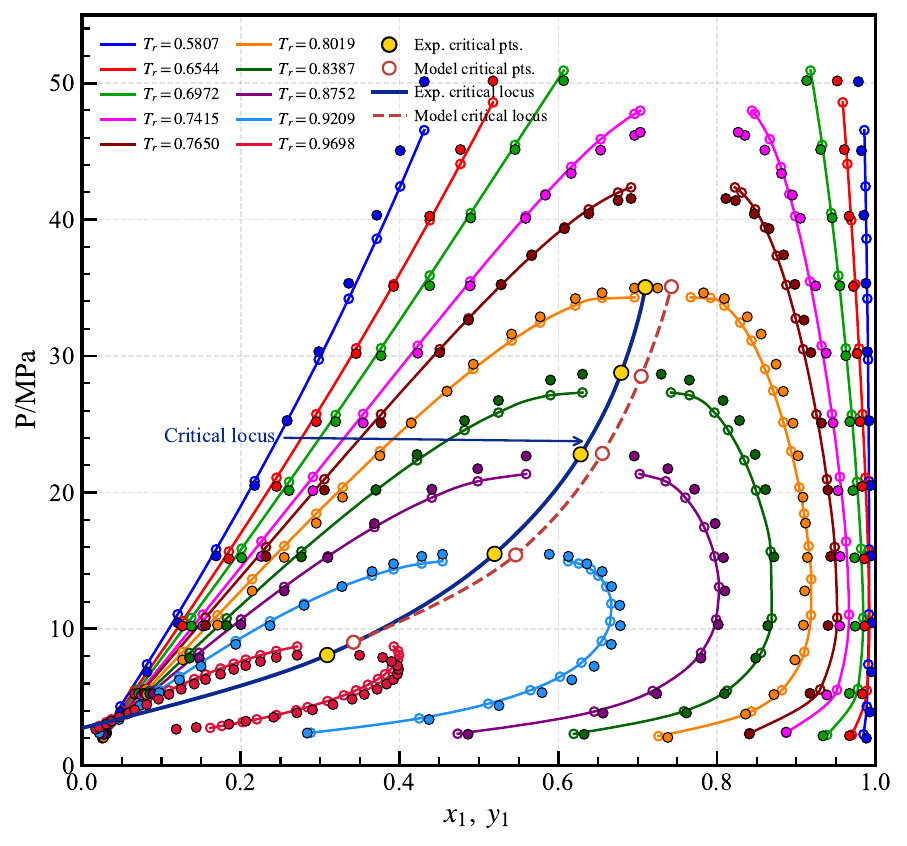}
\end{subfigure} 
\caption{
Representative pressure--composition phase diagram for the C$_7$H$_{16}$/N$_2$ system. The experimentally estimated critical points and the corresponding reference critical locus are included for visualization. The left panel shows the original PR-EOS prediction, whereas the right panel shows the symbolically corrected PR-EOS prediction.
}
\label{fig:critical_locus}
\end{figure}

As a representative example, Fig.~\ref{fig:critical_locus} presents the pressure--composition phase diagram for the C$_7$H$_{16}$/N$_2$ system together with the experimentally estimated critical points and a reference critical locus. The reference critical locus is included solely to facilitate visualization of the overall phase behavior. Compared with the original PR-EOS, the symbolically corrected PR-EOS more closely follows the experimental phase envelope while preserving the overall thermodynamic trends. Improved agreement is also observed in the vicinity of the estimated critical region, further demonstrating that the proposed symbolic correction enhances the predictive capability of the PR-EOS without altering its underlying thermodynamic structure.

\section{Conclusion}

In this work, we proposed a multilevel symbolic regression framework for correcting the Peng--Robinson equation of state for hydrocarbon--nitrogen vapor--liquid equilibrium prediction. System-specific symbolic correction terms were first identified from experimental data, and recurring symbolic structures were subsequently organized into a shared symbolic basis with coefficient functions parameterized by the carbon number. The resulting model provides compact and interpretable symbolic correction expressions while improving prediction accuracy for both pressure and vapor composition across multiple hydrocarbon--nitrogen systems. Numerical results demonstrated that the proposed two-level symbolic regression consistently outperformed both the original PR-EOS and a conventional symbolic regression model in terms of prediction accuracy. The proposed correction also produced improved agreement with the experimental pressure--composition diagrams and representative phase envelopes while preserving the overall thermodynamic behavior of the underlying PR-EOS.

The present study considered symbolic corrections to the PR-EOS for hydrocarbon--nitrogen mixtures. Future work will investigate equation-level symbolic corrections that directly modify the analytical form of the equation of state while preserving physical consistency.

\label{sec:conclusion}

\bibliographystyle{unsrt}
\bibliography{mybib}

@article{Reitz1987,
  author  = {Reitz, R. D.},
  title   = {Modeling Atomization Processes in High-Pressure Vaporizing Sprays},
  journal = {Atomization and Spray Technology},
  volume  = {3},
  pages   = {309--337},
  year    = {1987}
}

@article{ReitzBeale1999,
  author  = {Reitz, R. D. and Beale, J. C.},
  title   = {Modeling Spray Atomization With The Kelvin--Helmholtz/Rayleigh--Taylor Hybrid Model},
  journal = {Atomization and Sprays},
  volume  = {9},
  pages   = {623--650},
  year    = {1999},
  doi     = {10.1615/atomizspr.v9.i6.40}
}

@article{Qiu2015,
  author  = {Qiu, L. and Reitz, R. D.},
  title   = {An Investigation of Thermodynamic States During High-Pressure Fuel Injection using Equilibrium Thermodynamics},
  journal = {International Journal of Multiphase Flow},
  volume  = {72},
  pages   = {24--38},
  year    = {2015},
  doi     = {10.1016/j.ijmultiphaseflow.2015.01.011}
}

@incollection{QiaoHighPressure,
  author    = {Qiao, L. and Jain, S. and Mo, G.},
  title     = {Molecular Simulations to Research Supercritical Fuel Properties},
  booktitle = {High Pressure Flows for Propulsion Applications},
  editor    = {Bellan, J.}
}

@article{GarciaSanchez2009,
  author  = {Garc{\'i}a-S{\'a}nchez, F. and Eliosa-Jim{\'e}nez, G. and Silva-Oliver, G. and Garc{\'i}a-Flores, B. E.},
  title   = {Vapor-liquid Equilibrium Data for the Nitrogen + n-Decane System from (344 to 563) K and at Pressures up to 50 MPa},
  journal = {Journal of Chemical and Engineering Data},
  volume  = {54},
  pages   = {1560--1568},
  year    = {2009},
  doi     = {10.1021/je800881t}
}

@article{Privat2008a,
  author  = {Privat, R. and Jaubert, J.-N. and Mutelet, F.},
  title   = {Addition of the Nitrogen Group to the PPR78 Model},
  journal = {Industrial \& Engineering Chemistry Research},
  volume  = {47},
  pages   = {2033--2048},
  year    = {2008}
}

@article{Privat2008b,
  author  = {Privat, R. and Jaubert, J.-N. and Mutelet, F.},
  title   = {Use of the PPR78 Model to Predict New Equilibrium Data of Binary Systems Involving Hydrocarbons and Nitrogen},
  journal = {Industrial \& Engineering Chemistry Research},
  volume  = {47},
  pages   = {7483--7489},
  year    = {2008},
  doi     = {10.1021/ie800636h}
}

@article{Azarnoosh1963,
  author  = {Azarnoosh, A. and McKetta, J. J.},
  title   = {Nitrogen--n-Decane System in the Two-Phase Region},
  journal = {Journal of Chemical and Engineering Data},
  volume  = {8},
  pages   = {494--496},
  year    = {1963},
  doi     = {10.1021/je60019a005}
}

@article{Llave1988,
  author  = {Llave, F. M. and Chung, T. H.},
  title   = {Vapor-Liquid Equilibria of Nitrogen-Hydrocarbon Systems at Elevated Pressures},
  journal = {Journal of Chemical and Engineering Data},
  volume  = {33},
  pages   = {123--128},
  year    = {1988},
  doi     = {10.1021/je00052a019}
}

@article{Tong1999,
  author  = {Tong, J. and Gao, W. and Robinson, R. L. and Gasem, K. A. M.},
  title   = {Solubilities of Nitrogen in Heavy Normal Paraffins from 323 to 423 K at Pressures to 18.0 MPa},
  journal = {Journal of Chemical and Engineering Data},
  volume  = {44},
  pages   = {784--787},
  year    = {1999},
  doi     = {10.1021/je980279n}
}

@article{DAvila1976,
  author  = {D'Avila, S. G. and Kaul, B. K. and Prausnitz, J. M.},
  title   = {Solubilities of Heavy Hydrocarbons in Compressed Methane and Nitrogen},
  journal = {Journal of Chemical and Engineering Data},
  volume  = {21},
  pages   = {488--491},
  year    = {1976},
  doi     = {10.1021/je60071a017}
}

@article{Prausnitz1959,
  author  = {Prausnitz, J. M. and Benson, P. R.},
  title   = {Solubility of Liquids in Compressed Hydrogen, Nitrogen, and Carbon Dioxide},
  journal = {AIChE Journal},
  volume  = {5},
  year    = {1959}
}

@article{Pearce1993,
  author  = {Pearce, D. L. and Peters, C. J. and de Swaan Arons, J.},
  title   = {Measurement of the Gas Phase Solubility of Decane in Nitrogen},
  journal = {Fluid Phase Equilibria},
  volume  = {89},
  pages   = {335--343},
  year    = {1993},
  doi     = {10.1016/0378-3812(93)85092-Z}
}

@article{Gao1999,
  author  = {Gao, W. and Robinson, R. L. and Gasem, K. A. M.},
  title   = {High-pressure Solubilities of Hydrogen, Nitrogen, and Carbon Monoxide in Dodecane},
  journal = {Journal of Chemical and Engineering Data},
  volume  = {44},
  pages   = {130--132},
  year    = {1999},
  doi     = {10.1021/je9801664}
}

@article{GarciaCordova2011,
  author  = {Garcia-Cordova, F. and Justo-Garc{\'i}a, D. N. and Garc{\'i}a-Flores, B. E. and Garc{\'i}a-S{\'a}nchez, F.},
  title   = {Vapor-Liquid Equilibrium Data for the Nitrogen and Dodecane System at Temperatures from (344 to 593) K and at Pressures up to 60 MPa},
  journal = {Journal of Chemical and Engineering Data},
  volume  = {56},
  pages   = {1555--1564},
  year    = {2011},
  doi     = {10.1021/je1012372}
}

@article{PengRobinson1976,
  author  = {Peng, D. Y. and Robinson, D. B.},
  title   = {A New Two-Constant Equation of State},
  journal = {Industrial \& Engineering Chemistry Fundamentals},
  volume  = {15},
  pages   = {59--64},
  year    = {1976},
  doi     = {10.1021/i160057a011}
}

@article{Lopez2017,
  author  = {Lopez-Echeverry, J. S. and Reif-Acherman, S. and Araujo-Lopez, E.},
  title   = {Peng--Robinson Equation of State: 40 Years Through Cubics},
  journal = {Fluid Phase Equilibria},
  volume  = {447},
  pages   = {39--71},
  year    = {2017}
}

@article{Fateen2013,
  author  = {Fateen, S. E. K. and Khalil, M. M. and Elnabawy, A. O.},
  title   = {Semi-empirical Correlation for Binary Interaction Parameters of the Peng--Robinson Equation of State},
  journal = {Journal of Advanced Research},
  volume  = {4},
  pages   = {137--145},
  year    = {2013},
  doi     = {10.1016/j.jare.2012.03.004}
}

@article{Mohammed2018,
  author  = {Mohammed, F. and Qasim, M. and Elamir, A. and Darwish, N. A.},
  title   = {Generalized Binary Interaction Parameters for Hydrogen--Heavy-n-Alkane Systems Using Peng--Robinson Equation of State},
  journal = {Chemical Engineering Communications},
  volume  = {205},
  pages   = {1226--1238},
  year    = {2018},
  doi     = {10.1080/00986445.2018.1442333}
}

@article{Abudour2014,
  author  = {Abudour, A. M. and Mohammad, S. A. and Robinson, R. L. and Gasem, K. A. M.},
  title   = {Generalized Binary Interaction Parameters for the Peng--Robinson Equation of State},
  journal = {Fluid Phase Equilibria},
  volume  = {383},
  pages   = {156--173},
  year    = {2014},
  doi     = {10.1016/j.fluid.2014.10.006}
}

@article{Mohanty2005,
  author  = {Mohanty, S.},
  title   = {Estimation of Vapour Liquid Equilibria of Binary Systems Using Artificial Neural Networks},
  journal = {Fluid Phase Equilibria},
  volume  = {235},
  pages   = {92--98},
  year    = {2005},
  doi     = {10.1016/j.fluid.2005.07.003}
}

@inproceedings{Petersen1994,
  author    = {Petersen, R. and Fredenslund, A. and Rasmussen, P.},
  title     = {Artificial Neural Networks as a Predictive Tool for Vapor-Liquid Equilibrium},
  booktitle = {Computers \& Chemical Engineering},
  year      = {1994}
}

@article{Sharma1999,
  author  = {Sharma, R. and Singhal, D. and Ghosh, R. and Dwivedi, A.},
  title   = {Potential Applications of Artificial Neural Networks to Thermodynamics: Vapor-Liquid Equilibrium Predictions},
  journal = {Computers \& Chemical Engineering},
  volume  = {23},
  pages   = {385--390},
  year    = {1999},
  doi     = {10.1016/S0098-1354(98)00281-6}
}

@article{Ganguly2003,
  author  = {Ganguly, S.},
  title   = {Prediction of VLE Data Using Radial Basis Function Network},
  journal = {Computers \& Chemical Engineering},
  volume  = {27},
  pages   = {1445--1454},
  year    = {2003},
  doi     = {10.1016/S0098-1354(03)00068-1}
}

@article{Mohanty2006,
  author  = {Mohanty, S.},
  title   = {Estimation of Vapour Liquid Equilibria for the System Carbon Dioxide-Difluoromethane Using Artificial Neural Networks},
  journal = {International Journal of Refrigeration},
  volume  = {29},
  pages   = {243--249},
  year    = {2006},
  doi     = {10.1016/j.ijrefrig.2005.05.007}
}

@article{Ghanadzadeh2008,
  author  = {Ghanadzadeh, H. and Ahmadifar, H.},
  title   = {Estimation of Vapour + Liquid Equilibrium of Binary Systems Using an Artificial Neural Network},
  journal = {Journal of Chemical Thermodynamics},
  volume  = {40},
  pages   = {1152--1156},
  year    = {2008},
  doi     = {10.1016/j.jct.2008.02.011}
}

@article{Habiballah1996,
  author  = {Habiballah, W. A. and Startzman, R. A. and Barrufet, M. A.},
  title   = {Use of Neural Networks for Prediction of Vapor/Liquid Equilibrium K-Values for Light-Hydrocarbon Mixtures},
  journal = {SPE Reservoir Engineering},
  volume  = {11},
  pages   = {121--126},
  year    = {1996},
  doi     = {10.2118/28597-PA}
}

@article{Abedini2011,
  author  = {Abedini, R. and Zanganeh, I. and Mohagheghian, M.},
  title   = {Simulation and Estimation of Vapor-Liquid Equilibrium for Asymmetric Binary Systems Using Artificial Neural Network},
  journal = {Journal of Phase Equilibria and Diffusion},
  volume  = {32},
  pages   = {105--114},
  year    = {2011},
  doi     = {10.1007/s11669-011-9851-8}
}

@article{Wang2018,
  author  = {Xingjian Wang and Hongfa Huo and Umesh Unnikrishnan and Vigor Yang},
  title   = {A Systematic Approach to High-Fidelity Modeling and Efficient Simulation of Supercritical Fluid Mixing and Combustion},
  journal = {Combustion and Flame},
  volume  = {196},
  pages   = {364--375},
  year    = {2018},
  doi     = {10.1016/j.combustflame.2018.04.030}
}

@article{Privat2013,
  author  = {R. Privat and J.-N. Jaubert},
  title   = {Are Cubic Equations of State Still Suitable Tools for the Correlation and Prediction of the Thermodynamic Properties of Pure Compounds?},
  journal = {Journal of Chemical Thermodynamics},
  volume  = {63},
  pages   = {139--154},
  year    = {2013},
  doi     = {10.1016/j.jct.2013.03.012}
}

@book{Koza1992,
  author    = {John R. Koza},
  title     = {Genetic Programming: On the Programming of Computers by Means of Natural Selection},
  publisher = {MIT Press},
  year      = {1992}
}

@article{Schmidt2009,
  author  = {Michael Schmidt and Hod Lipson},
  title   = {Distilling Free-Form Natural Laws from Experimental Data},
  journal = {Science},
  volume  = {324},
  number  = {5923},
  pages   = {81--85},
  year    = {2009},
  doi     = {10.1126/science.1165893}
}

@article{chakraborty2022vapor,
  title={Vapor--liquid equilibrium estimation of n-alkane/nitrogen mixtures using neural networks},
  author={Chakraborty, Suman and Sun, Yixuan and Lin, Guang and Qiao, Li},
  journal={Journal of Computational and Applied Mathematics},
  volume={408},
  pages={114059},
  year={2022},
  publisher={Elsevier}
}

@article{Cranmer2023,
  author  = {Miles Cranmer},
  title   = {Interpretable Machine Learning for Science with PySR and SymbolicRegression.jl},
  journal = {arXiv preprint arXiv:2305.01582},
  year    = {2023}
}

@article{Rumelhart1986,
  author  = {David E. Rumelhart and Geoffrey E. Hinton and Ronald J. Williams},
  title   = {Learning Representations by Back-Propagating Errors},
  journal = {Nature},
  volume  = {323},
  number  = {6088},
  pages   = {533--536},
  year    = {1986},
  doi     = {10.1038/323533a0}
}

@article{SilvaOliver2006,
  author  = {Guadalupe Silva-Oliver and Gaudencio Eliosa-Jim{\'e}nez and Fernando Garc{\'i}a-S{\'a}nchez and Juan R. Avenda{\~n}o-G{\'o}mez},
  title   = {High-pressure vapor--liquid equilibria in the nitrogen--n-pentane system},
  journal = {Fluid Phase Equilib.},
  volume  = {250},
  number  = {1--2},
  pages   = {37--48},
  year    = {2006},
  doi     = {10.1016/j.fluid.2006.09.018}
}

@article{EliosaJimenez2007,
  author  = {Gaudencio Eliosa-Jim{\'e}nez and Guadalupe Silva-Oliver and Fernando Garc{\'i}a-S{\'a}nchez and Antonio de Ita de la Torre},
  title   = {High-Pressure Vapor--Liquid Equilibria in the Nitrogen + n-Hexane System},
  journal = {J. Chem. Eng. Data},
  volume  = {52},
  number  = {2},
  pages   = {395--404},
  year    = {2007},
  doi     = {10.1021/je060341d}
}

@article{GarciaSanchez2007,
  author  = {Fernando Garc{\'i}a-S{\'a}nchez and Gaudencio Eliosa-Jim{\'e}nez and Guadalupe Silva-Oliver and Armando God{\'i}nez-Silva},
  title   = {High-pressure (vapor + liquid) equilibria in the (nitrogen + n-heptane) system},
  journal = {J. Chem. Thermodyn.},
  volume  = {39},
  number  = {6},
  pages   = {893--905},
  year    = {2007},
  doi     = {10.1016/j.jct.2006.11.007}
}

@article{SilvaOliver2007,
  author  = {Guadalupe Silva-Oliver and Gaudencio Eliosa-Jim{\'e}nez and Fernando Garc{\'i}a-S{\'a}nchez and Juan Ram{\'o}n Avenda{\~n}o-G{\'o}mez},
  title   = {High-pressure vapor--liquid equilibria in the nitrogen--n-nonane system},
  journal = {J. Supercrit. Fluids},
  volume  = {42},
  number  = {1},
  pages   = {36--47},
  year    = {2007},
  doi     = {10.1016/j.supflu.2007.01.006}
}

@article{yang2025symbolic,
  title={Symbolic-regression aided development of a new cubic equation of state for improved liquid phase density calculation at pressures up to 100 MPa},
  author={Yang, Xiaoxian and Frotscher, Ophelia and Richter, Markus},
  journal={International Journal of Thermophysics},
  volume={46},
  number={2},
  pages={29},
  year={2025},
  publisher={Springer}
}

\appendix
\section{}

This appendix summarizes the optimized parameters of the proposed multilevel symbolic regression model. Tables~\ref{tab:multilevel_dp_basis_constants} and~\ref{tab:multilevel_dy_basis_constants} list the learned internal constants associated with the shared symbolic basis functions for the pressure and vapor-composition corrections, respectively. These constants define the functional forms of the shared basis functions and are optimized using the merged training dataset.

\begin{table}[htp!]
\centering
\scriptsize
\caption{Learned internal constants for the shared pressure basis functions.}
\label{tab:multilevel_dp_basis_constants}
\renewcommand{\arraystretch}{1.15}
\begin{tabular}{c l l}
\hline
Basis & Parameters & Values \\
\hline
$\phi_1^{(P)}$ & $(a_1^{(P)},a_2^{(P)},a_3^{(P)})$ &
$(0.714,\ 1.60,\ -1.30)$ \\
$\phi_2^{(P)}$ & $(b_1^{(P)},b_2^{(P)},b_3^{(P)})$ &
$(-0.397,\ 2.01,\ 0.449)$ \\
$\phi_3^{(P)}$ & $(c_1^{(P)},c_2^{(P)},c_3^{(P)})$ &
$(-0.656,\ 2.13,\ 0.649)$ \\
$\phi_4^{(P)}$ & $(d_1^{(P)},d_2^{(P)},d_3^{(P)},d_4^{(P)},d_5^{(P)},d_6^{(P)})$ &
$(0.739,\ -0.388,\ -0.173,\ 0.052,\ -0.086,\ 0.316)$ \\
$\phi_5^{(P)}$ & $(e_1^{(P)},e_2^{(P)})$ &
$(1.88,\ 2.31)$ \\
\hline
\end{tabular}
\end{table}

\begin{table}[htp!]
\centering
\scriptsize
\caption{Learned internal constants for the shared vapor-composition basis functions.}
\label{tab:multilevel_dy_basis_constants}
\renewcommand{\arraystretch}{1.15}
\begin{tabular}{c l l}
\hline
Basis & Parameters & Values \\
\hline
$\phi_1^{(y)}$ & $(a_1^{(y)},a_2^{(y)},a_3^{(y)})$ &
$(0.610,\ 0.219,\ 0.696)$ \\
$\phi_2^{(y)}$ & $(b_1^{(y)},b_2^{(y)})$ &
$(1.09,\ 0.443)$ \\
$\phi_3^{(y)}$ & $(c_1^{(y)},c_2^{(y)},c_3^{(y)})$ &
$(0.672,\ 0.914,\ 0.254)$ \\
$\phi_4^{(y)}$ & $(d_1^{(y)},d_2^{(y)},d_3^{(y)})$ &
$(1.34,\ 2.42,\ 0.127)$ \\
$\phi_5^{(y)}$ & $(e_1^{(y)},e_2^{(y)},e_3^{(y)},e_4^{(y)},e_5^{(y)},e_6^{(y)})$ &
$(0.605,\ 0.279,\ 1.10,\ 0.561,\ -0.217,\ -0.028)$ \\
$\phi_6^{(y)}$ & $(g_1^{(y)},g_2^{(y)})$ &
$(-0.353,\ 1.16)$ \\
\hline
\end{tabular}
\end{table}

Tables~\ref{tab:multilevel_dp_coefficients} and~\ref{tab:multilevel_dy_coefficients} report the system-specific coefficients associated with each shared basis function. These coefficients determine the contribution of each basis function to the correction model for individual hydrocarbon--nitrogen systems and constitute the reference data used for the coefficient interpolation described in Section~3.
For completeness and reproducibility, all optimized basis constants and system-specific coefficients are provided explicitly. Together with the symbolic basis functions presented in the main text, these parameters fully specify the proposed multilevel symbolic regression model and enable direct reconstruction of the pressure and vapor-composition correction functions.

\begin{table}[htp!]
\centering
\scriptsize
\caption{System-specific coefficients $\alpha_m^{(P,s)}$ for the multilevel pressure model $\Delta P^{(s)}(T_r,x_{N_2})=\sum_{m=1}^{5}\alpha_m^{(P,s)}\phi_m^{(P)}(T_r,x_{N_2})$.}
\label{tab:multilevel_dp_coefficients}
\renewcommand{\arraystretch}{1.15}
\begin{tabular}{lccccc}
\hline
System & $\alpha_1^{(P,s)}$ & $\alpha_2^{(P,s)}$ & $\alpha_3^{(P,s)}$ & $\alpha_4^{(P,s)}$ & $\alpha_5^{(P,s)}$ \\
\hline
C$_5$/N$_2$ & $-0.039$ & $0.358$ & $0.144$ & $0.216$ & $-0.052$ \\
C$_6$/N$_2$ & $-0.095$ & $0.348$ & $0.148$ & $0.260$ & $0.003$ \\
C$_7$/N$_2$ & $-0.176$ & $0.356$ & $0.158$ & $0.331$ & $0.014$ \\
C$_9$/N$_2$ & $-0.342$ & $0.175$ & $0.387$ & $0.437$ & $-0.084$ \\
C$_{10}$/N$_2$ & $-0.328$ & $0.245$ & $0.107$ & $0.657$ & $0.055$ \\
C$_{12}$/N$_2$ & $-0.435$ & $0.328$ & $0.221$ & $0.676$ & $-0.165$ \\
\hline
\end{tabular}
\end{table}

\begin{table}[htp!]
\centering
\scriptsize
\caption{System-specific coefficients $\alpha_m^{(y,s)}$ for the multilevel vapor-composition model $\Delta y^{(s)}(T_r,x_{N_2})=\sum_{m=1}^{6}\alpha_m^{(y,s)}\phi_m^{(y)}(T_r,x_{N_2})$.}
\label{tab:multilevel_dy_coefficients}
\renewcommand{\arraystretch}{1.15}
\begin{tabular}{lcccccc}
\hline
System & $\alpha_1^{(y,s)}$ & $\alpha_2^{(y,s)}$ & $\alpha_3^{(y,s)}$ & $\alpha_4^{(y,s)}$ & $\alpha_5^{(y,s)}$ & $\alpha_6^{(y,s)}$ \\
\hline
C$_5$/N$_2$ & $0.122$ & $0.199$ & $0.151$ & $1.11$ & $-1.03$ & $0.379$ \\
C$_6$/N$_2$ & $0.094$ & $0.192$ & $0.077$ & $1.15$ & $-1.01$ & $0.387$ \\
C$_7$/N$_2$ & $0.077$ & $0.178$ & $0.003$ & $1.13$ & $-0.895$ & $0.390$ \\
C$_9$/N$_2$ & $0.074$ & $-0.032$ & $-0.008$ & $0.622$ & $-0.038$ & $0.315$ \\
C$_{10}$/N$_2$ & $0.022$ & $-0.052$ & $-0.013$ & $0.259$ & $0.191$ & $0.256$ \\
C$_{12}$/N$_2$ & $0.233$ & $0.141$ & $-0.124$ & $-0.189$ & $0.448$ & $0.195$ \\
\hline
\end{tabular}
\end{table}

\section{Explicit forms of the multilevel correction model}
\label{app:explicit_multilevel_equations}

This appendix summarizes the explicit form of the multilevel correction model used in the main text.
The model is constructed from shared symbolic basis functions together with system-specific coefficients.
For each hydrocarbon--nitrogen system, the pressure correction and the vapor-composition correction are represented as linear combinations of the corresponding shared basis functions.
The corrected predictions are then obtained by adding these correction terms to the original PR-EOS predictions.

For notational simplicity, we write
\[
T \equiv T_r,
\qquad
x \equiv x_{N_2}.
\]
With this notation, the corrected pressure and vapor composition for system $s$ are given by
\[
P_{\mathrm{corr}}^{(s)}(T,x)=P_{\mathrm{PR}}+\Delta P^{(s)}(T,x),
\qquad
y_{\mathrm{corr}}^{(s)}(T,x)=y_{\mathrm{PR}}+\Delta y^{(s)}(T,x).
\]

The pressure correction is expressed as
\[
\Delta P^{(s)}(T,x)=\sum_{m=1}^{5}\alpha_m^{(P,s)}\phi_m^{(P)}(T,x),
\]
where $\phi_m^{(P)}$ denote the shared basis functions for pressure and $\alpha_m^{(P,s)}$ are the system-specific coefficients.
Similarly, the vapor-composition correction is expressed as
\[
\Delta y^{(s)}(T,x)=\sum_{m=1}^{6}\alpha_m^{(y,s)}\phi_m^{(y)}(T,x),
\]
where $\phi_m^{(y)}$ denote the shared basis functions for vapor composition and $\alpha_m^{(y,s)}$ are the corresponding system-specific coefficients.

\subsection{Shared basis functions for \texorpdfstring{$\Delta P$}{Delta P}}

The shared pressure basis functions are
\begin{align}
\phi_1^{(P)}(T,x)
&=
0.714\left(x+1.60-e^{-1.30T+x}\right), \\
\phi_2^{(P)}(T,x)
&=
-0.397\,(T+x^2)(x+2.01)+0.449, \\
\phi_3^{(P)}(T,x)
&=
-0.656\,(T+x-2.13)(x+0.649), \\
\phi_4^{(P)}(T,x)
&=
\left(0.739T-0.388x\right)
\left(e^{\,x-0.173(T+0.052x^2)}+0.086\right)+0.316, \\
\phi_5^{(P)}(T,x)
&=
\frac{x-e^x+1.88}{(2.31)^T}.
\end{align}

Using these basis functions, the system-specific pressure corrections are written as
\begin{align}
\Delta P^{(\mathrm{C5})}(T,x)
&=
-0.039\,\phi_1^{(P)}
+0.358\,\phi_2^{(P)}
+0.144\,\phi_3^{(P)}
+0.216\,\phi_4^{(P)}
-0.052\,\phi_5^{(P)}, \\
\Delta P^{(\mathrm{C6})}(T,x)
&=
-0.095\,\phi_1^{(P)}
+0.348\,\phi_2^{(P)}
+0.148\,\phi_3^{(P)}
+0.260\,\phi_4^{(P)}
+0.003\,\phi_5^{(P)}, \\
\Delta P^{(\mathrm{C7})}(T,x)
&=
-0.176\,\phi_1^{(P)}
+0.356\,\phi_2^{(P)}
+0.158\,\phi_3^{(P)}
+0.331\,\phi_4^{(P)}
+0.014\,\phi_5^{(P)}, \\
\Delta P^{(\mathrm{C9})}(T,x)
&=
-0.342\,\phi_1^{(P)}
+0.175\,\phi_2^{(P)}
+0.387\,\phi_3^{(P)}
+0.437\,\phi_4^{(P)}
-0.084\,\phi_5^{(P)}, \\
\Delta P^{(\mathrm{C10})}(T,x)
&=
-0.328\,\phi_1^{(P)}
+0.245\,\phi_2^{(P)}
+0.107\,\phi_3^{(P)}
+0.657\,\phi_4^{(P)}
+0.055\,\phi_5^{(P)}, \\
\Delta P^{(\mathrm{C12})}(T,x)
&=
-0.435\,\phi_1^{(P)}
+0.328\,\phi_2^{(P)}
+0.221\,\phi_3^{(P)}
+0.676\,\phi_4^{(P)}
-0.165\,\phi_5^{(P)}.
\end{align}

\subsection{Shared basis functions for \texorpdfstring{$\Delta y$}{Delta y}}

The shared basis functions for the vapor-composition correction are
\begin{align}
\phi_1^{(y)}(T,x)
&=
-x\left(0.610+e^T(0.219x+0.696)\right), \\
\phi_2^{(y)}(T,x)
&=
1.09-x-e^{0.443Tx}, \\
\phi_3^{(y)}(T,x)
&=
(0.672T-0.914)(x+0.254T^2), \\
\phi_4^{(y)}(T,x)
&=
(T+1.34)\left(\frac{x}{T-2.42}+0.127\right), \\
\phi_5^{(y)}(T,x)
&=
\left(0.605-0.279(x-e^x)e^{1.10T}\right)
\left(0.561-0.217xe^T\right)+0.028, \\
\phi_6^{(y)}(T,x)
&=
-0.353\,x\,(e^T)^{(1.16)^x}.
\end{align}

The corresponding system-specific vapor-composition corrections are
\begin{align}
\Delta y^{(\mathrm{C5})}(T,x)
&=
0.122\,\phi_1^{(y)}
+0.199\,\phi_2^{(y)}
+0.151\,\phi_3^{(y)}
+1.11\,\phi_4^{(y)}
-1.03\,\phi_5^{(y)}
+0.379\,\phi_6^{(y)}, \\
\Delta y^{(\mathrm{C6})}(T,x)
&=
0.094\,\phi_1^{(y)}
+0.192\,\phi_2^{(y)}
+0.077\,\phi_3^{(y)}
+1.15\,\phi_4^{(y)}
-1.01\,\phi_5^{(y)}
+0.387\,\phi_6^{(y)}, \\
\Delta y^{(\mathrm{C7})}(T,x)
&=
0.077\,\phi_1^{(y)}
+0.178\,\phi_2^{(y)}
+0.003\,\phi_3^{(y)}
+1.13\,\phi_4^{(y)}
-0.895\,\phi_5^{(y)}
+0.390\,\phi_6^{(y)}, \\
\Delta y^{(\mathrm{C9})}(T,x)
&=
0.074\,\phi_1^{(y)}
-0.032\,\phi_2^{(y)}
-0.008\,\phi_3^{(y)}
+0.622\,\phi_4^{(y)}
-0.038\,\phi_5^{(y)}
+0.315\,\phi_6^{(y)}, \\
\Delta y^{(\mathrm{C10})}(T,x)
&=
0.022\,\phi_1^{(y)}
-0.052\,\phi_2^{(y)}
-0.013\,\phi_3^{(y)}
+0.259\,\phi_4^{(y)}
+0.191\,\phi_5^{(y)}
+0.256\,\phi_6^{(y)}, \\
\Delta y^{(\mathrm{C12})}(T,x)
&=
0.233\,\phi_1^{(y)}
+0.141\,\phi_2^{(y)}
-0.124\,\phi_3^{(y)}
-0.189\,\phi_4^{(y)}
+0.448\,\phi_5^{(y)}
+0.195\,\phi_6^{(y)}.
\end{align}

\end{document}